\documentclass{article}

\usepackage{amsfonts,amsmath,bm,booktabs}
\usepackage{amsthm}

\theoremstyle{plain}
\newtheorem{theo}{Theorem}
\newtheorem{lemm}[theo]{Lemma}
\newtheorem{rema}[theo]{Remark}

\newcommand{\eq}[1]{(\ref{#1})}

\newcommand{\RR}{\mathbb{R}}

\newcommand{\bE}{\mathbb{E}}
\newcommand{\bN}{\mathbb{N}}

\newcommand{\cE}{{\cal E}}
\newcommand{\cF}{{\cal F}}
\newcommand{\cT}{{\cal T}}

\usepackage{times}
\usepackage{graphicx} 
\usepackage{subfigure} 

\usepackage{natbib}

\newcommand{\argmax}{\mathop{\rm arg~max}}

\usepackage{hyperref}

\usepackage[accepted]{icml2018}

\icmltitlerunning{Selective Inference for Change Point Detection in Multi-dimensional Sequences}

\begin{document} 

\twocolumn[
\icmltitle{Selective Inference for Change Point Detection in Multi-dimensional Sequences}

\begin{icmlauthorlist}
\icmlauthor{Yuta Umezu}{nitech}
\icmlauthor{Ichiro Takeuchi}{nitech,riken,nims} 
\end{icmlauthorlist}

\icmlaffiliation{nitech}{Nagoya Institute of Technology, Nagoya, Japan}
\icmlaffiliation{riken}{RIKEN, Tokyo, Japan}
\icmlaffiliation{nims}{National Institute for Materials Science, Tokyo, Japan}

\icmlcorrespondingauthor{Ichiro Takeuchi}{takeuchi.ichiro@nitech.ac.jp}

\icmlkeywords{Change Point Detection, Selective Inference}

\vskip 0.3in
]

\printAffiliationsAndNotice{}  

\begin{abstract} 
We study the problem of detecting change points (CPs) that are characterized by a subset of dimensions in a multi-dimensional sequence. 
A method for detecting those CPs can be formulated as a two-stage method: one for selecting relevant dimensions, and another for selecting CPs.
It has been difficult to properly control the false detection probability of these CP detection methods because selection bias in each stage must be properly corrected.
Our main contribution in this paper is to formulate a CP detection problem as a selective inference problem, and show that exact (non-asymptotic) inference is possible for a class of CP detection methods. 
We demonstrate the performances of the proposed selective inference framework through numerical simulations and its application to our motivating medical data analysis problem. 
\end{abstract} 

\section{Introduction}
\label{sec:intro}
\begin{figure*}[t!]
 \centering
 \vskip -.6375in
 \includegraphics[scale=.61, viewport=0 0 720 306]{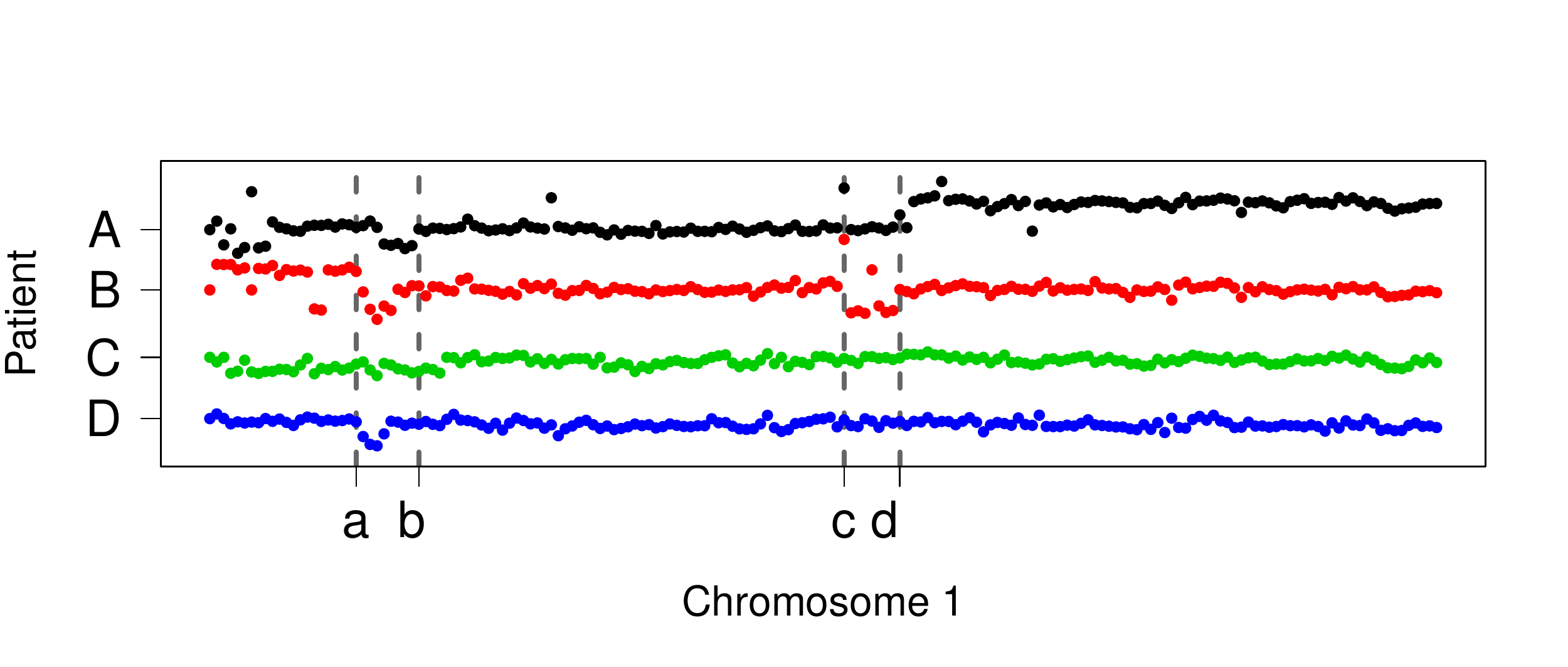}
 \vskip -.2in
 \caption{
 A motivating example taken from a biomedical study~\citep{takeuchi2009potential}. 
 The four-dimensional sequence indicates the results of copy number variations of four malignant lymphoma patients in Chromosome 1.
 The positive/negative values in each dimension indicate the copy number gain/loss at the corresponding genome position. 
 The purpose of this biomedical study is to detect common change points (CPs) in a subset of dimensions which indicates that the copy number variations are relevant to the malignant lymphoma disease. 
 The difficulty lies in the fact that copy number variations are heterogeneous among patients. 
 For example, patients B and D have a common CP at position a, while patients A and B share a CP at position b. 
 In this paper, we propose a class of methods that can detect common CPs shared by a subset of dimensions in a multi-variate sequence. 
 In \S\ref{sec:simulation}, we analyze this malignant lymphoma dataset in order to demonstrate the effectiveness of the proposed methods. 
 }
 \label{fig:aCGH}
 \vskip -.2in
\end{figure*}

In this paper, we study change point (CP) detection in multi-dimensional sequences.
This problem arises in many areas. 
Figure~\ref{fig:aCGH} shows a motivating example in which copy number variations of four malignant lymphoma patients are plotted. 
The purpose of this study is to detect common CPs among patients.
The difficulty of this problem lies in the fact that the copy number variations are heterogeneous among patients, i.e., they are commonly observed only in a subset of patients as illustrated in the figure. 

The goal of this paper is to develop a method for detecting CPs that are characterized by a subset of dimensions in a multi-dimensional sequence. 
A CP detection method for solving such a problem can be formulated as a two-stage method: one for selecting dimensions, and another for selecting time points\footnote{Although our target is not restricted to \emph{time} sequences, we use the term ``time'' to refer to a point in a sequence.}. 
We call the former the aggregation stage, and the latter the scanning stage.
In the aggregation stage, a subset of dimensions is selected and the scores for the selected dimensions are aggregated into a scalar score that represents the plausibility of a common change at the time point. 
Various forms of aggregation can be considered.
In the scanning stage, a time-point that maximizes the aggregated score is selected as a CP by scanning the one-dimensional aggregated score sequence. 

In this paper, we our goal is not only to detect CPs but also to properly control the false detection probability at the desired significance level, e.g., $\alpha = 0.05$. 
To this end, we must take into account the fact that the CPs are obtained by \emph{selecting} the dimensions and the time points in the aggregation and scanning stages, respectively. 
This means that two types of \emph{selection bias} must be properly corrected in order to make inferences on the detected CPs. 
For a one-dimensional sequence, various statistical CP detection methods have been studied~\citep{Pag54, yao1988estimating, lee2003cusum, yau2016inference}.
In contrast, for multi-dimensional sequences, there are only a few asymptotic results and these were developed under rather restrictive assumptions. 
Deriving the sampling distributions of CP statistics is generally difficult even in asymptotic scenarios because stochastic singularity of CPs must be properly handled. 

In this paper, we introduce a class of CP detection methods that can exactly control the false detection probability. 
Our key contribution is to interpret a CP detection problem as a \emph{selective inference} problem. 
For the class of CP detection methods, this interpretation allows us to derive the exact (non-asymptotic) sampling distribution of the test CP statistic conditional on the fact that the CP is selected based on a particular method. 
Using a recent result on selective inference~\citep{taylor2015statistical, Lee16}, we show that inferences based on the derived sampling distribution can correct the two types of selection bias in the aggregation and scanning stages, and the overall false detection probability can be properly controlled. 

The rest of the paper is organized as follows.
In \S\ref{sec:setup_relatedworks}, we formulate our problem setup and review related work. 
Here, we focus on a problem of detecting a single mean structure change in an off-line manner. 
In \S\ref{sec:propose}, we introduce a selective inference interpretation of a CP detection problem, and present the main results. 
Here, we first introduce a class of CP detection methods which includes many existing ones. 
We then show that the \emph{selective type I error} of the detected CPs can be exactly (non-asymptotically) controlled. 
We also show that our hypothesis testing procedure is an \emph{approximately unbiased test} and derive a lower bound of the power in the sense of selective inference.
In \S\ref{sec:extension}, the results in the previous section are extended to multiple CP detection problems via a local hypothesis testing framework. 
\S\ref{sec:simulation} is devoted to numerical experiments.
Here, we first confirm that the proposed CP detection methods can properly control the false detection probabilities in a simulation study.
We then apply these CP detection methods to copy number variation analysis of 46 malignant lymphoma patients. 

\paragraph{Notations}
We define $\RR_+=\{x\in\RR\mid x\ge 0\}$, $[n, m] := \{n, \ldots, m\}$ for $n, m\in \bN$ and, as a special case, $[n]:=[1, n]$.
For two matrices $P$ and $Q$, $\otimes$ denotes the Kronecker product, i.e., $P\otimes Q=(P_{ij}Q)_{ij}$.
We use notation $\|\bm{v}\|_P^2=\bm{v}^\top P\bm{v}$ for an appropriate matrix $P$ and vector $\bm{v}$.
For a non-negative integer $N$, an $N$-by-$N$ identity matrix is denoted by $I_N$, while a zero matrix is denoted as $O$, omitting its size as long as no confusion is expected.
The sign function, the indicator function and the vectorize operator are denoted by $\text{sgn}(\cdot), I\{\cdot\}$ and $\text{vec}(\cdot)$, respectively.

\section{Problem Setup and Related Work}
\label{sec:setup_relatedworks}
In this section, we present the problem setup and discuss related works.
We first consider the problem of detecting a single CP.
Its extensions to multiple CP detection is discussed in \S\ref{sec:extension}. 

\subsection{Problem Setup}
Let us write 
an $N$-dimensional sequence with length $T$
as an $N$-by-$K$ matrix
$Y=(\bm{y}_1,\ldots, \bm{y}_T)  \in \RR^{N \times K}$.
Then, a single CP detection problem for mean shift is formulated as the following hypothesis testing problem
\begin{align}
\nonumber
 &\text{H}_0:\bE[\bm{y}_t]=\bE[\bm{y}_{t+1}],
~~~
{}^\forall t\in[T-1]
~~~~~
\text{vs.} \\
\label{eq:test_all}
 &\text{H}_1:\bE[\bm{y}_t]\neq \bE[\bm{y}_{t+1}],
~~~
{}^{\exists !} t\in[T-1],
\end{align}
where 
the null hypothesis $\text{H}_0$ states that the mean vector does not change within the entire sequence, 
whereas 
the alternative hypothesis $\text{H}_1$ states that there is one CP.

\paragraph{CUSUM score}
In order to discuss the test statistic and its sampling distribution for the hypothesis testing problem in \eq{eq:test_all},
let us simply consider a single CP detection from a one-dimensional sequence denoted as $\bm{y}=(y_1,\ldots, y_T)^\top$.
Let $Y_t$ and $Z_t$ be generic random variables corresponding to $\{y_1,\ldots, y_t\}$ and $\{y_{t+1},\ldots, y_T\}$, respectively.
Then, we expect the time point $t$ to be a CP when the value $|\bE[Y_t]-\bE[Z_t]|$ is large.
Hence, we define a natural estimator of discrepancy between random variables $Y_t$ and $Z_t$ as
\begin{align*}
\left|
\frac{1}{t}\sum_{u=1}^t y_u-\frac{1}{T-t}\sum_{u=t+1}^T y_u
\right|,
\end{align*}
and, its scaled measure
\begin{align*}
S(t)
=\left\{\frac{t(T-t)}{T}\right\}^{1/2}\left(\frac{1}{t}\sum_{u=1}^t y_u -\frac{1}{T-t}\sum_{u=t+1}^T y_u \right)
\end{align*}
is known as the \emph{CUSUM (cumulative sum) score} \citep{Pag54}.
Note that the CUSUM score can be interpreted as a realization of the logarithm of a Gaussian likelihood when we assume that the sequence $\bm{y}$ is mutually independent.
A point that maximizes $|S(t)|$ is detected as a CP.

\paragraph{Multi-variate CUSUM score and its aggregation}
In the case of a multi-dimensional sequence, it is natural to consider a multi-variate version of the CUSUM score: 
\begin{align}
\label{eq:CUSUM}
 \bm S(t) := (S_1(t), \ldots, S_N(t))^\top \in \RR^N
\end{align}
for $t \in [T-1]$, where each $S_i(t)$ is a CUSUM score corresponding to the $i$-th dimension.
Since each element of a multi-variate CUSUM score cannot be maximized simultaneously, we need to first \emph{aggregate} the $N$-dimensional vector $\bm S(t)$ into a scalar value. 
We denote an aggregation function as $\cF: \RR^N \to \RR_+$, i.e., $\bm{S}(t)\mapsto {\cal F}(\bm{S}(t))$ for each $t\in[T-1]$.
The aggregated score $\cF(\bm{S}(t))$ represents the plausibility of a CP at time point $t$.

\paragraph{Choices of aggregation function}
For multi-dimensional CP detection, various choices of aggregation function $\cF$ can be considered. 

\subsubsection*{$\ell_\infty$-aggregation}
\citet{Jir15} proposed $\ell_\infty$-aggregation as
$\max_{i\in[N]}|S_i(t)|$.
This aggregation function simply selects the dimension whose absolute CUSUM score is greatest among the $N$ dimensions.
This choice is not appropriate when there are changes in multiple dimensions.

\subsubsection*{$\ell_1$-aggregation}
Another simple aggregation function is $\ell_1$-aggregation defined as
$\sum_{i\in[N]}|S_i(t)|$.
This aggregation function just sums up the individual CUSUM scores.
This choice is not appropriate if changes are observed only in a subset of dimensions.

\subsubsection*{Top $K$-aggregation}
If changes are observed in a subset of dimensions and the size of the subset is known to be $K$, then the top $K$-aggregation function defined as
$\max_{I \subseteq [N] : |I| = K}\sum_{i \in I} |S_i(t)|$ is appropriate.
This aggregation function can be interpreted as a generalization of $\ell_\infty$- and $\ell_1$-aggregation functions since it reduces to $\ell_\infty$- and $\ell_1$-aggregation when $K=1$ and $K=N$, respectively. 

\subsubsection*{Double CUSUM aggregation}
If $K$ is unknown, then it would be nice to be able select the appropriate $K$ from the data. 
Let $\rho_j(t)$, $j \in [N]$ be the $j$-th largest value in $\{|S_i(t)|\}_{i \in [N]}$, i.e., $\rho_1(t) \ge \cdots \ge \rho_N(t)$ is satisfied for $t \in [T-1]$. 
\citet{Cho16} proposed the double CUSUM aggregation function defined as
\begin{align}
\label{eq:DC}
\max_{k\in [N-1]}\gamma_k^{\varphi}\left(\frac{1}{k}\sum_{i=1}^k\rho_i(t)-\frac{1}{2N-k}\sum_{i=k+1}^N\rho_i(t)\right),
\end{align}
where $\gamma_k=k(2N-k)/(2N)$ and $\varphi$ is a pre-determined positive constant.
This aggregation function returns the CUSUM score of the sequence $\rho_1(t), \ldots, \rho_N(t)$ for each $t\in[T-1]$.
The rationale for this choice is that, if there are changes in $k$ dimensions, then the top $k$ absolute CUSUM values $\{\rho_1(t),\ldots, \rho_k(t)\}$ tend to have larger values than the remaining CUSUM values $\{\rho_{k+1}(t),\ldots, \rho_N(t)\}$, meaning that the CUSUM score for this sequence would be maximized at $k$. 
As suggested in \citet{Cho16}, $\varphi=0.5$ would be optimal in the sense of asymptotic theory.

\paragraph{Test statistic for the problem in \eq{eq:test_all}}
Based on the above discussion, a natural test statistic for the hypothesis testing in \eq{eq:test_all} is
\begin{align}
\label{eq:test-statistic}
\theta
=\max_{t \in [T-1]} \cF(\bm S(t)).
\end{align}
Here, $\theta$ can be interpreted as a realization of the corresponding random variable $\hat{\theta}$, and thus we could consider $\hat{\theta}$ as a test statistic.
Then, the $p$-value is defined as the false detection probability under $\text{H}_0$, that is, $\text{P}(\hat{\theta}\ge \theta)$.
If the $p$-value is smaller than the significance level $\alpha$, then we can conclude that there is a single CP in the multi-dimensional sequence. 

As we briefly discuss in the following subsection, 
it is difficult (even asymptotically) to derive the sampling distribution of $\hat{\theta}$ in~\eq{eq:test-statistic} for most practical choices of the aggregation function $\cF$ unless rather restrictive assumptions are imposed. 

\subsection{Related Work}
Here, we briefly review existing work on controlling the false detection probabilities in CP detection problems. 

\paragraph{Inference for one-dimensional sequences}
First, we review statistical inferences on CP detection in one-dimensional sequences.
As described above, one can regard that the point $t$ at which the CUSUM score $|S(t)|$ maximized is the most plausible as a CP.
Hence, the test statistic of the target hypothesis is naturally defined as $\max_{t\in[T-1]}|S(t)|$, and corresponds to the log-likelihood ratio test statistic.
Then, as is well known, test statistic $\max_{t\in[T-1]}|S(t)|/\sigma$ converges weakly to a Brownian bridge under appropriate moment and weak dependence assumptions on the sequence \citep{Phi87, csorgo1997limit, Sha10}, 
where $\sigma^2=\lim_{T\to \infty} T\text{Var}[\sum_{t \in [T] }y_t/T]$ is the so-called long-run variance. 
In this asymptotic theory, the weak dependence assumption is essential because, otherwise, the long-run variance does not exist or becomes zero. 

Another closely related existing inference procedure is proposed in \citet{hyun2016exact}. 
They used fused LASSO~\citep{Tib05} for CP detection, and interpret the inferences on the detected CPs as the inferences on the regression model coefficients for the selected features by fused LASSO. 
Inferences on the regression model coefficients for selected features can be done using a recently popularized framework on selective inference~\cite{Lee16}. 

Although inference problems on CP detection in one-dimensional sequences have been intensively studied in the literature, they cannot be easily generalized to the case of multi-dimensional sequences. 

\paragraph{Inference for multi-dimensional sequences}
Unlike the one-dimensional sequence case, the literature of inference on multi-dimensional sequences is very scarce\footnote{Several methods for assuring estimation were developed \citep{fryzlewicz2014wild, cho2015multiple, Cho16, wang2016high}, but they cannot be used for inference on the detected CPs.}.
To the best of our knowledge, existing work on this topic can be defined into two types.

\begin{table}[t!]
\vskip -0.2in
\caption{Example WRAG functions.}
\label{tab:score}
\vskip 0.05in
\begin{center}
\begin{small}
\begin{sc}
\begin{tabular}{ll}
\toprule
method & weight: $c_{k,i}$ \\
\midrule
$\ell_1$-aggregation & $1$ \\[.05in]
$\ell_\infty$-aggregation & $I\{i=1\}$ \\[.05in]
top $K$-aggregation & $I\{i\leq K\}$ \\[.05in]
double CUSUM &\hskip-.1in $\begin{array}{ll} \alpha_k^\varphi/kI\{i\leq k\} \\[.03in] -\alpha_k^\varphi/(2N-k)I\{i\geq k+1\} \end{array}$ \\
\bottomrule
\end{tabular}
\end{sc}
\end{small}
\end{center}
\vskip -0.2in
\end{table}

The first type studies likelihood-based methods such as $\ell_\infty$-aggregation reported by \citet{Jir15}.
They derived an asymptotic distribution of $\ell_\infty$-aggregation score 
$\max_{t \in [T-1]} \max_{i \in [N]} |S_i(t)|$
as an extreme value distribution. 
To establish the asymptotic distribution, 
the relation between the length of the sequence $T$ and the size of the dimension $N$ must satisfy 
$T=O(N^\delta)$
for some
$\delta\in(0,1)$.
This condition indicates that 
if $N$ is relatively large compared to  $T$,
then one can no longer control the false detection probability even when the underlying distribution is independent for each time point. 

The second type uses a kernel-based method. 
The basic idea of kernel-based CP detection is
to consider the problem
as a two-sample test
where the two multi-dimensional subsequences before and after the CP
are regarded as the two samples. 
In this approach,
some discrepancy measure between the two samples
such as
the kernel Fisher discriminant ratio
or
maximum mean discrepancy (MMD) measures 
is defined, 
and 
then
the test statistic is the maximum value of the discrepancy measure scanned along the sequence.
\citet{harchaoui2009kernel} first studied kernel CP detection by using the kernel Fisher discriminant ratio, while \citet{li2015m} employed MMD as the discrepancy measure. 
They derived an asymptotic distribution of the test statistic (i.e., the maximum discrepancy along the sequence) under the assumption that values at different time points are independently distributed. 

\section{Selective Inference for Multi-dimensional CP Detection}
\label{sec:propose}

In this section, we present our main results.
As discussed in the previous section, it is difficult to derive the sampling distribution of the test statistic $\theta$ in the form of \eq{eq:test-statistic}.
Our basic idea to overcome this difficulty is to interpret the CP detection problem as a selective inference problem, i.e., the problem of making an inference on the detected CP conditional on the fact that the CP is selected by a particular choice of aggregation function. 
This interpretation enables us to derive the exact (non-asymptotic) \emph{selective} sampling distribution of the test statistic for a wide class of practical aggregation functions $\cF$. 

\subsection{Proposed Class of Aggregation Functions}
Let us first propose a class of aggregation functions for which we can derive the exact selective sampling distribution of the test statistic. 
Recall that $S_i(t)$ is the CUSUM score in the $i$-th dimension at the $t$-th time point for $i \in [N], t \in [T-1]$.
As in the definition of the double CUSUM aggregation function, $\rho_j(t)$, $j \in [N]$ is defined as the $j$-th largest value in $\{|S_j(t)|\}_{i \in [N]}$, i.e., $\rho_1(t) \ge \cdots \ge \rho_N(t)$ is satisfied for $t \in [T-1]$. 
We define a class of aggregation functions as
\begin{align}
\label{eq:WRAG}
\cF(\bm{S}(t))
=\max_{k\in [N-1]}\sum_{j=1}^Nc_{k,j}\rho_j(t), 
\end{align}
where $c_{k,j}, k \in [N-1], j \in [N]$ are constants. 
We refer to this class of aggregation functions as \emph{weighted rank aggregation (WRAG)} functions. 
Table \ref{tab:score} shows several choices of the constants $c_{k,j}$ corresponding to the choices of the aggregation functions discussed in \S\ref{sec:setup_relatedworks}. 
With the use of a WRAG function,
we can detect a CP $\hat{t}$ as 
\begin{align}
 \label{eq:hat_t}
 \hat{t}
 = \argmax_{t \in [T-1]} \max_{k\in [N-1]}\sum_{j=1}^Nc_{k,j}\rho_j(t).
\end{align}
We refer to a CP detection method via \eq{eq:hat_t} as a WRAG method. 

\subsection{Selective Inference on the CPs by WRAG methods}
An inference on the CP detected by a WRAG method can be interpreted as a selective inference, which has been actively studied in the past few years for inferences on feature selection problems~\citep{FitSunTay14, yang2016selective, tian2017asymptotics, suzumura2017selective}. 
In order to formulate our problem as a selective inference problem, let us define the \emph{selection event} and the \emph{selected test statistic}. 
The selection event in a WRAG method is written as 
\begin{align*}
 \{(\hat{t}, \hat{k}) = (t, k)\}, 
\end{align*}
where
$t$ and $k$
are interpreted as realizations of the corresponding random variables 
$\hat{t}$ 
and
$\hat{k}$
defined as 
\begin{align*}
(\hat{t}, \hat{k})
= \argmax_{(t,k) \in [T-1]\times [N-1]}\sum_{j=1}^Nc_{k,j}\rho_j(t).
\end{align*}
On the other hand, the selected test statistic is given as 
\begin{align}
 \label{eq:selected-test-statistic}
 \theta_{k}(t) = \sum_{j=1}^N c_{k, j} \rho_j(t).
\end{align}
In the context of selective inference, based on the selection event and the selected test statistic, the so-called \emph{selective $p$-value} is defined as
\begin{align*}
 \text{P}
 \bigl(
 \hat{\theta}_{k}(t) > \theta_k(t)
 \mid
 (\hat{t}, \hat{k}) = (t, k)
 \bigr)
\end{align*}
where $\theta_k(t)$ is interpreted as a realization of the corresponding statistic $\hat{\theta}_{k}(t)$. 

For the purpose of inference, we make an assumption on the normality of multi-dimensional sequence $Y$, namely,
\begin{align*}
 {\rm vec}(Y) \sim \text{N}_{NT}({\rm vec}(M), \Xi \otimes \Sigma), 
\end{align*}
where
$M \in \RR^{N \times T}$ is the mean matrix, 
whereas 
$\Xi \in \RR^{T \times T}$
and
$\Sigma \in \RR^{N \times N}$
are covariance matrices
representing
the time correlation
and
the variable correlation
structures,
respectively.
In practical CP detection tasks, it is often possible to obtain sequences without CPs, i.e., samples from the null hypothesis.
The covariance matrices $\Xi$ and $\Sigma$ can be estimated from such samples or manually specified based on prior knowledge.
Denoting the mean matrix as $M = (\bm m_1, \ldots, \bm m_T)$, the null hypothesis in our selective inference is written as
\begin{align}
 \label{eq:selected_null}
 \text{H}_0: \bm m_{\hat{t}} = \bm m_{\hat{t}+1},
\end{align}
where, remember, $\hat{t}$ is the detected CP via a WRAG method, meaning that it is a \emph{random} variable. 

In order to derive the sampling distribution of the selected test statistic, we use a recent seminal result in \citet{Lee16}. 
We first slightly generalize the key lemma in their work for handling a random matrix $Y$. 
\begin{lemm}[Polyhedral Lemma for A Random Matrix]
\label{lem1}
Suppose that ${\rm vec}(Y) \sim \text{N}_{NT}({\rm vec}(M), \Xi \otimes \Sigma)$ with mean ${\rm vec}(M)$ and covariance $\Xi \otimes \Sigma$. 
Let $\bm{\gamma}=(\Xi\bm{\eta}\otimes\Sigma\bm{\delta})(\|\bm{\eta}\|_\Xi^2\|\bm{\delta}\|_\Sigma^2)^{-1}$ for any $\bm{\delta}\in\mathbb{R}^N$ and $\bm{\eta}\in\mathbb{R}^T$, and let $\bm{z}=(I_{NT}-\bm{\gamma}(\bm{\eta}\otimes\bm{\delta})^\top){\rm vec}(Y)$.
Then, any event represented in the form of $\{A {\rm vec}(Y) \le \bm b\}$  for a fixed matrix $A$ and a fixed vector $\bm b$ can be written as 
\begin{align*}
 \{A{\rm vec}(Y)\leq \bm{b}\} =\{L(\bm{z})\leq \bm{\delta}^\top Y\bm{\eta}\leq U(\bm{z}),\;N(\bm{z})\geq 0\},
\end{align*}
where
\begin{align}
\label{eq:truncation_points}
L(\bm{z})
=\max_{l:(A\bm{\gamma})_l<0}\frac{b_l-(A\bm{z})_l}{(A\bm{c})_l},
U(\bm{z})
=\min_{l:(A\bm{\gamma})_l>0}\frac{b_l-(A\bm{z})_l}{(A\bm{c})_l}
\end{align}
and $N(\bm{z})=\max_{l:(A\bm{c})_l=0}b_l-(A\bm{z})_l$.
In addition, $\bm{\delta}^\top Y\bm{\eta}$ is independent of $(L(\bm{z}),U(\bm{z}),N(\bm{z}))$.
\end{lemm}
The proof of the lemma is presented in Appendix~\ref{subsec:proof-lem1}. 
This lemma states that
if the test statistic is expressed as a bi-linear function with the matrix $Y$ 
in the form of
$\bm \delta^\top Y \bm \eta$, 
and
the selection event can be expressed as an affine constraint in the form of
$\{ A {\rm vec}(Y) \le \bm b \}$, 
then the selected test statistic
$\hat{\theta}_k(t)$
is restricted to a certain interval. 
This lemma is a simple extension of Lemma~5.1 in \citet{Lee16}.

The selection event
$\{(\hat{t}, \hat{k}) = (t, k)\}$
cannot be written in the form of
$\{A {\rm vec}(Y) \le \bm b\}$. 
Here,
we consider the signs and the permutations of
$\{S_i(t)\}_{i \in [N]}$
for each
$t \in [T-1]$
as additional selection events. 
Let
$P_t=\text{diag}(\text{sgn}(S_i(t)))$
be an 
$N$-by-$N$
diagonal matrix
whose diagonal elements are the signs of $\{S_i(t)\}_{i \in [N]}$ for each $t$,
and
$G_t$ be an $N$-by-$N$ permutation matrix which maps
$(|S_i(t)|)_{i\in[N]}$ to $(\rho_j(t))_{j\in[N]}$.
The selection event is then formulated as
\begin{align*}
 \cE :=
 \bigl\{(\hat{t}, \hat{k}) = (t, k) \bigr\}
 \bigcap_{u \in [T-1]}  \bigl\{ (\hat{G}_u, \hat{P}_u) = (G_u, P_u) \bigr\},
\end{align*}
where $G_u, P_u$ are interpreted as realizations of the corresponding statistics $\hat{G}_u, \hat{P}_u$ for $u \in [T-1]$. 

The following theorem is the core of the selective $p$-value computation in our selective inference. 
\begin{theo}
\label{thm2}
Assume that the conditions of Lemma \ref{lem1} hold.
Then, there exist $L, U \in \mathbb{R}_+$ and $v^2\in\mathbb{R}_+$ such that 
 \begin{align}
  \label{eq:truncateNormal}
  \Bigl[F_{0,v^2}^{[L,U]}(\hat{\theta}_k(t))\mid \cE\Bigr] \sim {\rm Unif}(0,1)
 \end{align}
 under the null hypothesis \eq{eq:selected_null},
 where
 $F_{\mu,\sigma^2}^{[l, u]}(\cdot)$
 is the cumulative distribution function (c.d.f.) of normal distribution
 $\text{N}(\mu,\sigma^2)$
 truncated to the interval $[l,u]$.
\end{theo}
The complete proof of Theorem~\ref{thm2} is presented in Appendix~\ref{subsec:proof-thm2}, where we show that,
for any choice of aggregation function $\cF$ from the class in \eq{eq:WRAG}, 
\begin{itemize}
 \item the selection event $\cE$ can be written as an affine constraint event in the form of $\{A {\rm vec}(Y) \le \bm b\}$,
 \item the selected test static $\hat{\theta}_k(t)$ can be written as a bi-linear function with a matrix $Y$ in the form of $\bm \delta^\top Y \bm \eta$.
\end{itemize}
Then, by applying Lemma~\ref{lem1} and Theorem 5.2 in \citet{Lee16}, Theorem~\ref{thm2} can be proved.

As described in Appendix~\ref{sec:example}, 
for any choice of the aggregation function $\cF$ from the class in \eq{eq:WRAG}, 
the values of $L$, $U$, and $v^2$ in Theorem~\ref{thm2} can be computed. 
By using these values, the selective $p$-value can be computed as
\begin{align*}
 1 - F_{0, v^2}^{[L, U]}(\theta_k(t))
  = \frac{\Phi(U/v)-\Phi(\theta_k(t)/v)}{\Phi(U/v)-\Phi(L/v)}, 
\end{align*}
where
$\Phi(\cdot)$ is the c.d.f. of the standard normal distribution. 

\begin{rema}
When we do not need to select $k$ from the data, e.g., $\ell_1$-, $\ell_\infty$- and top $K$-aggregations, we do not need to consider the event $\{\hat{k}=k\}$ since the event is redundant for inference.
For the same reason, we also do not need to consider the some of the signs or/and the permutations depending on the choice of the aggregation function. 
Concrete examples of truncation points for each choice are described in Appendix \ref{sec:example}.
\end{rema}

\begin{rema}
We can establish the same result as Theorem \ref{thm2} even if all the signs and the permutations are considered. 
In this case, instead of a single interval, multiple intervals must be considered for all possible choices of the permutations and the signs. 
However, since the number of all possible combinations of the signs and the permutations is large, this is computationally intractable. 
\end{rema}

\subsection{Power Analysis}
Since $\bm{\delta}^\top Y\bm{\eta}\sim \text{N}(\bm{\delta}^\top M\bm{\eta}, \|\bm{\eta}\|_\Xi^2\|\bm{\delta}\|_\Sigma^2)$ for any $\bm{\delta}$ and $\bm{\eta}$, hypothesis (\ref{eq:selected_null}) can be viewed as
\begin{align*}
\text{H}_0: \bm{\delta}^\top M\bm{\eta}=0.
\end{align*}
In practice, since $\bm{\delta}$ and $\bm{\eta}$ are determined by a WRAG method, the hypothesis is also random.
Therefore, we consider an alternative hypothesis
\begin{align}
\label{eq:alternative}
\text{H}_1: \bm{\delta}^\top M\bm{\eta}=\mu>0,
\end{align}
which is a negation of the null since our aggregated score takes only positive values.
Under the alternative, the same argument as in Theorem \ref{thm2} indicates that
\begin{align*}
\Bigl[F_{\mu,v^2}^{[L,U]}(\hat{\theta}_k(t))\mid \cE\Bigr]
\sim{\rm Unif}(0,1).
\end{align*}
We now consider the power of the test in the situation that $\mu\to0$.
Note that this asymptotic scenario corresponds to the power analysis under the local alternative.

\begin{theo}
\label{thm3}
Let $z_\alpha$ be the upper $\alpha$-quantile of the null distribution.
Then, under the alternative (\ref{eq:alternative}), the power of the test is approximated as follows:
\begin{align*}
&{\rm P}\left(\hat{\theta}_k(t)\geq z_\alpha\mid \cE\right) \\
&=\alpha+\frac{\kappa}{v} \mu+\frac{\kappa}{v^2}\frac{\phi(U/v)-\phi(L/v)}{\Phi(U/v)-\Phi(L/v)} \mu^2+O(\mu^3) \\
&\geq \frac{3}{4}\alpha+ \frac{1}{4}\frac{\phi(z_\alpha/v)-\phi(U/v)}{\phi(L/v)-\phi(U/v)}+O(\mu^3),
\end{align*}
almost surely, where 
\begin{align*}
\kappa
=\frac{\phi(z_\alpha/v)-\{\phi(U/v)-\alpha(\phi(U/v)-\phi(L/v))\}}{\Phi(U/v)-\Phi(L/v)}
\leq 0
\end{align*}
and $\phi(\cdot)$ is the probability density function of the standard normal distribution.
\end{theo}
The proof of Theorem~\ref{thm3} is presented in Appendix \ref{subsec:proof-thm3}.
The theorem states that our selective inference procedure is an approximately unbiased test. Here, an unbiased test refers to a test in which the power becomes $\alpha$ at the boundary of the hypothesis, i.e., $\mu=0$. In addition, the test has a power of at least $3\alpha/4$ since the second term in the last inequality is always positive. Theorem \ref{thm3} suggests that there may exist better tests in terms of power.

\section{Extension to Multiple CP Detection}
\label{sec:extension}
In this section, we extend the selective inference framework for WRAG methods to be able to detect multiple CPs. 
To this end, we introduce a sliding window approach. 
Let $W_h(t)=[t-h+1, t+h]$ be a sliding window centered at $t$ with length $2h$ for each $t\in[h, T-h]$.
If we simply conducted single CP detection within each sliding window, too many CPs would be detected due to overlaps of multiple windows. 
To circumvent this issue, \citet{hao2013multiple} considered the so-called \emph{local hypothesis testing problem}. 
For each window $W_h(t), t \in [h, T-h]$, a local hypothesis test is defined as 
\begin{align*}
 &\text{H}_{0,t}: \bm{m}_u=\bm{m}_{u+1},
 ~~~
 {}^\forall u\in W_h(t)
 ~~~~~
 \text{vs.} \\
 &\text{H}_{1,t}: \bm{m}_t\neq \bm{m}_{t+1}.
\end{align*}
In this hypothesis test, even when the null hypothesis is rejected, unless there is a CP at the center of the window, the hypothesis itself is considered to be out of our interest. 
A natural estimate of the set of CPs by this approach would be 
\begin{align*}
\cT =\Bigl\{t\in[h,T-h]\mid t=\argmax_{u\in[t-h+1,t+h-1]}\cF(\bm{S}(u))\Bigr\}
\end{align*}
In the context of one-dimensional CP detection problems, \citet{yau2016inference} referred to this type of multiple CP estimates as \emph{local change point estimates}. 
In this approach, we can only consider $|\cT|$ hypotheses, which is usually a much smaller number than that of windows $T-2h+1$. 
\citet{yau2016inference} also discussed the choice of the window size $h$, and claimed that the choice of $h=\Omega(\log T)$ would be appropriate in an asymptotic sense.

\section{Numerical Experiments}
\label{sec:simulation}
Here, we confirm the performances of the proposed selective inference framework for WRAG methods through numerical experiments with both synthetic and real data.

\subsection{Experiments on Synthetic Data}
\label{subsec:synthetic}

\subsubsection{FPRs of selective and naive inferences}
First, we confirmed whether the false positive rates (FPRs) are properly controlled in the selective inference framework for WRAG methods with double CUSUM (DC) and top $K$-aggregation with $K = 1,5, 10, 15, 20$, where $K=1$ and $K=20$ correspond to $\ell_\infty$- and $\ell_1$-aggregations, respectively. 
The synthetic data with $N=20$ and $T=100$ were generated from normal distribution $\text{N}_{NT}({\rm vec}(O), \Xi \otimes \Sigma)$ where $\Xi=(\xi^{|i-j|})_{i,j\in[T]}$ and $\Sigma=(\sigma^{|i-j|})_{i,j\in[N]}$.
We considered $\sigma \in \{0.0, 0.5\}$ for simulating the cases without and with correlation among different dimensions, while $\xi$ was changed from 0.0 to 1.0 to simulate the cases with various degrees of correlation among different time points. 
In addition, we also computed the FPRs of naive inference for WRAG methods \emph{without} any selection bias correction procedure as in Theorem \ref{thm2}.
The significance level was set as $\alpha=0.05$. 
In all cases, 1,000 runs with different random seeds were simulated. 

\begin{figure}[t!]
\centering
\vskip -.275in
\subfigure[$\sigma=0$]{
\includegraphics[width=.5\linewidth, viewport=0 0 360 360]{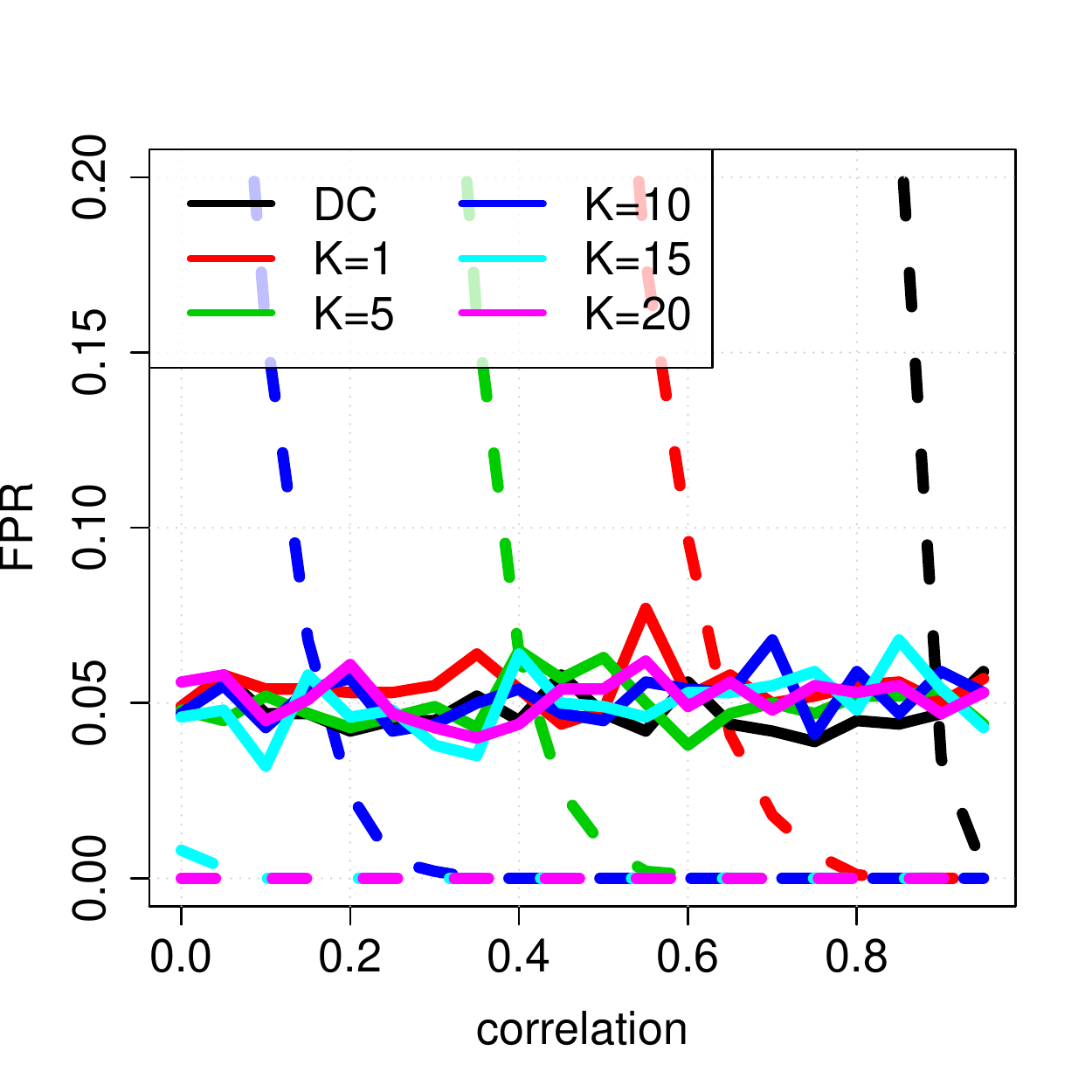}
\label{subfig:FPR0}
}
\hskip -.175in
\subfigure[$\sigma=0.5$]{
\includegraphics[width=.5\linewidth, viewport=0 0 360 360]{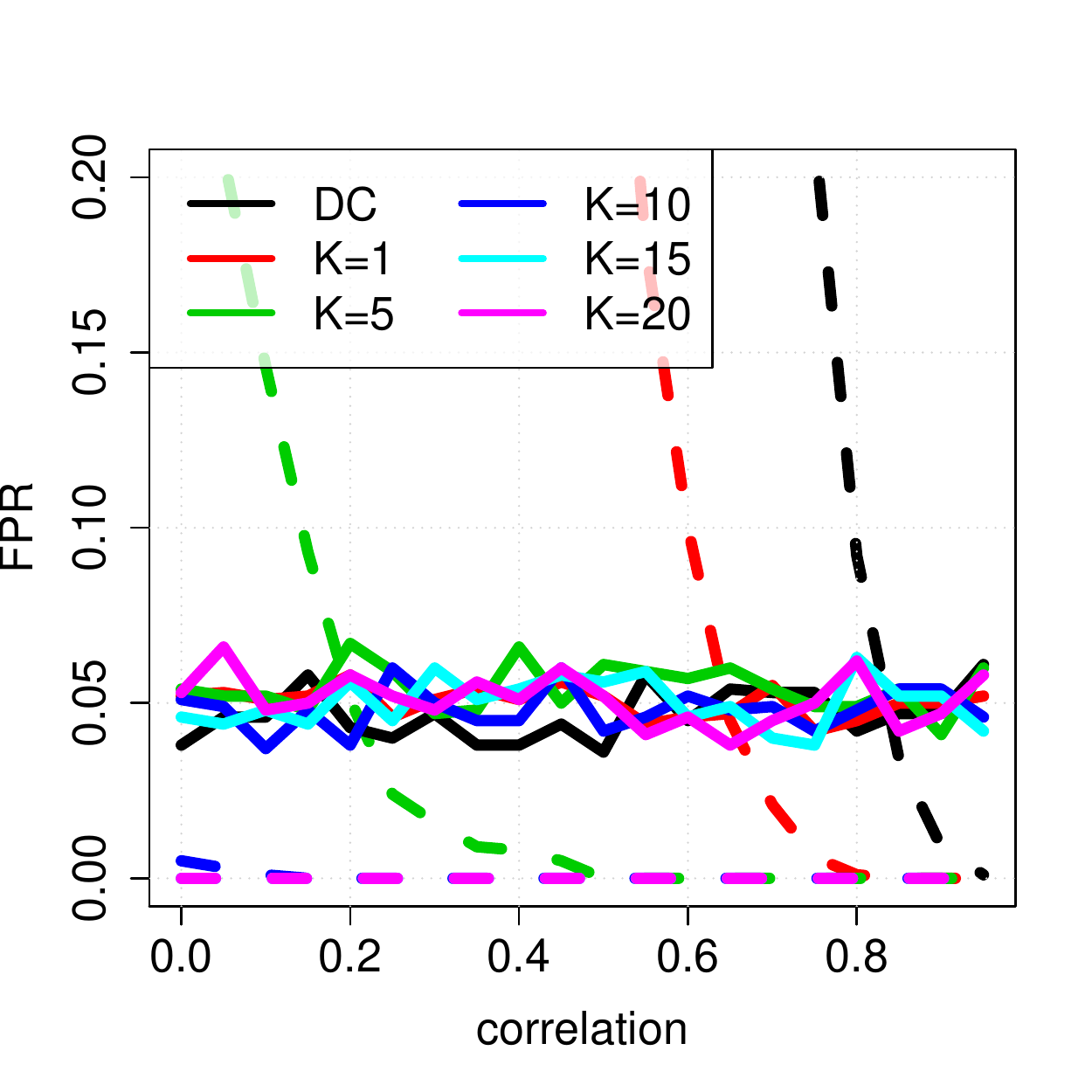}
\label{subfig:FPR0.5}
}
\vskip -.1in
\caption{
False positive rates (FPRs) of selective inference (solid) and naive inference (dashed).
The left and the right plots show the results without and with correlation among different dimensions.
In all cases, selective inference could properly control FPRs, while naive inference failed. 
}
\label{fig:FPR}
\vskip -.2in
\end{figure}

\begin{figure}[t!]
\centering
\vskip -.2in
\subfigure[selective $p$-value]{
\includegraphics[width=.5\linewidth, viewport=0 0 360 360]{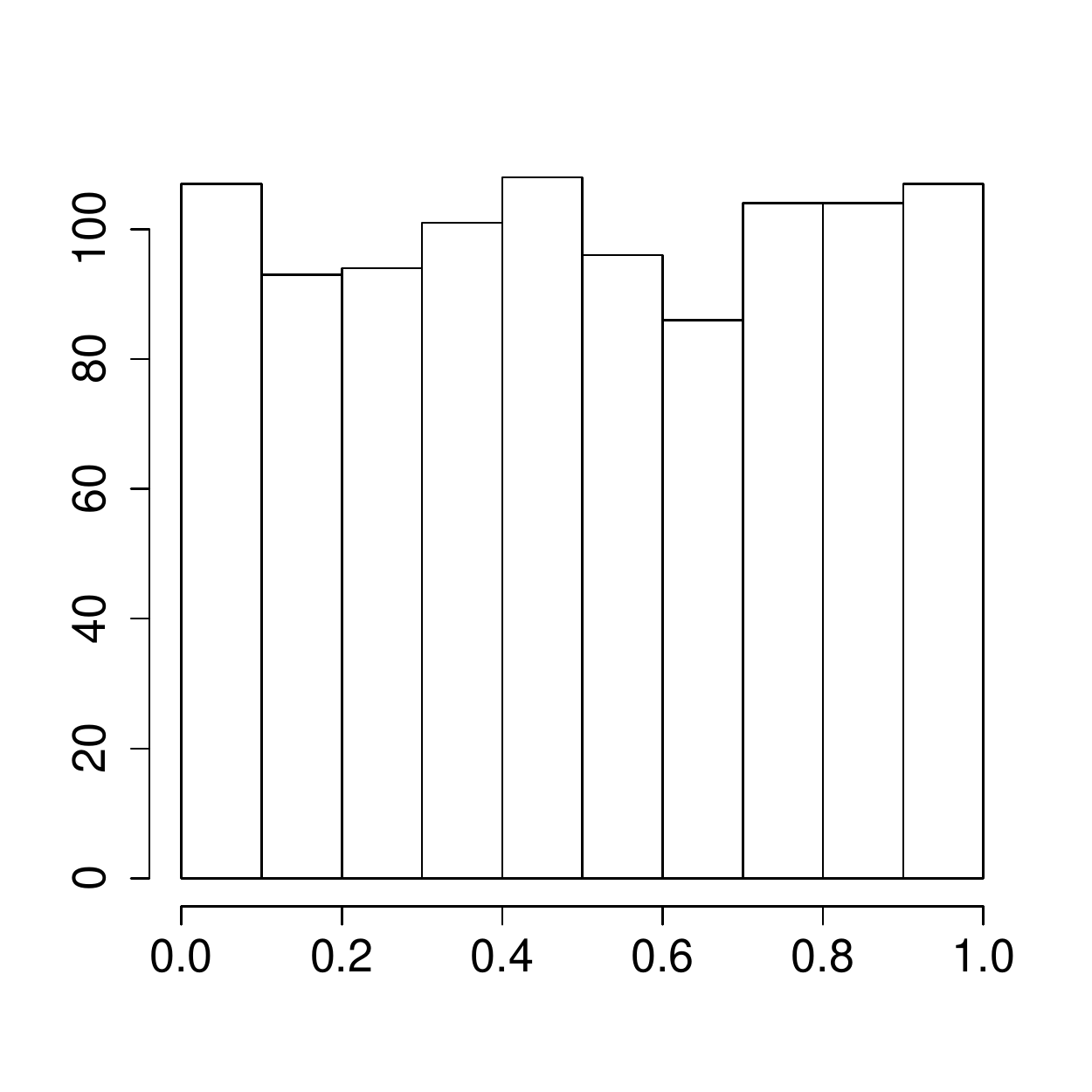}
\label{subfig:selective}
}
\hskip -.175in
\subfigure[naive $p$-value]{
\includegraphics[width=.5\linewidth, viewport=0 0 360 360]{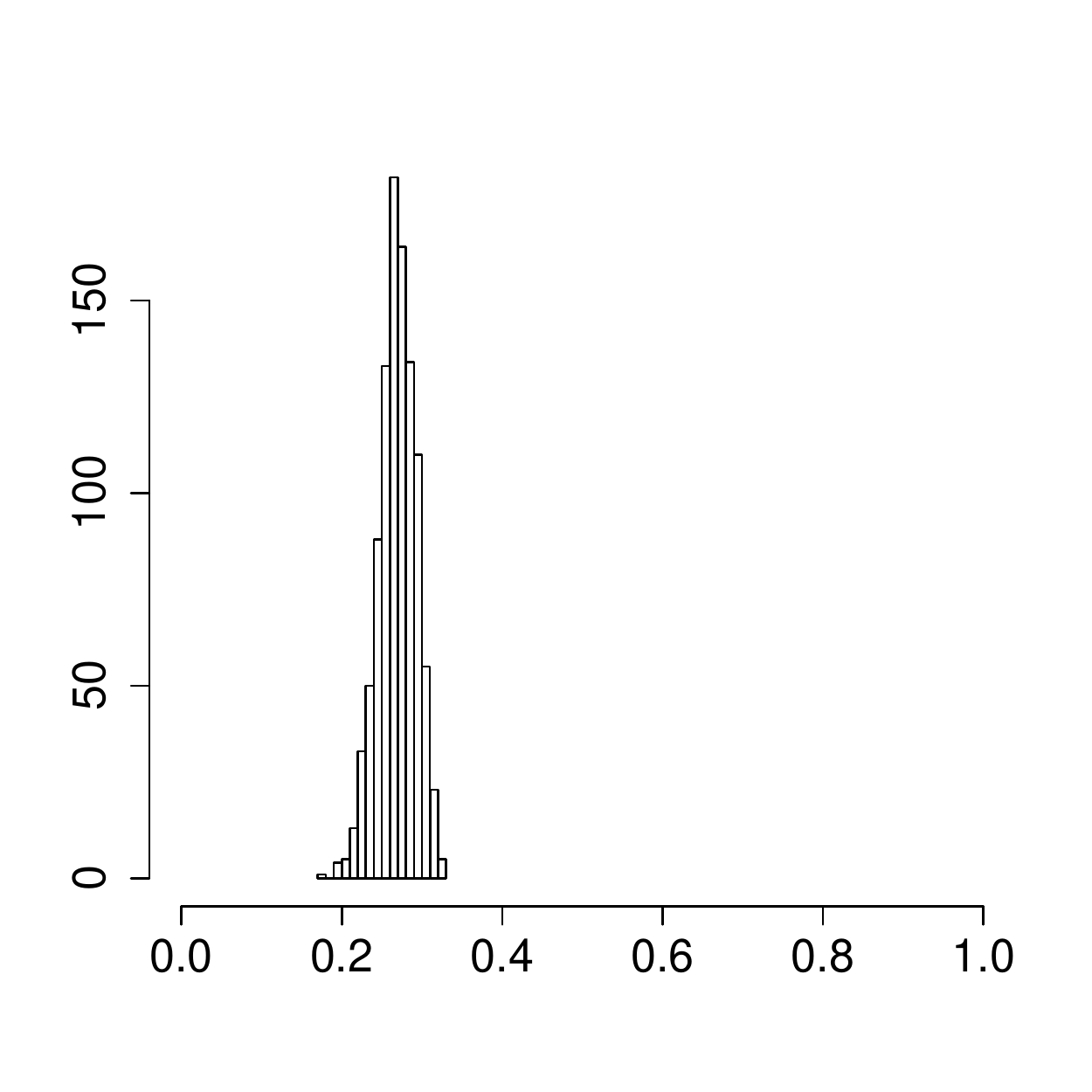}
\label{subfig:naive}
}
\vskip -.1in
\caption{
Histograms of (a) selective $p$-values and (b) naive $p$-values when $\xi=0.05$ and $\sigma=0.0$.
Noting that the $p$-values should be uniformly distributed between $0$ and $1$ under the null hypothesis, the left plot indicates that selective inference behaves as desired, while the right plot suggests that naive inference behaves incorrectly. 
}
\label{fig:histogram}
\vskip -.2in
\end{figure}

Figure \ref{fig:FPR} shows the FPRs of selective inference (solid lines) and naive inference (dashed lines), where the horizontal and vertical axes indicate the value of $\xi$ and the estimated FPRs, respectively.
We see that selective inference could control FPRs appropriately in all cases. 
On the other hand, in almost all cases, naive inference  failed to control the FPRs especially when $\xi$ is small\footnote{The bias of naive inference is large when the ``effective'' length of the sequence is large. Since effective length decreases as the degree of correlation increases, the bias is large when $\xi$ is small.}.
Although the results in Figure \ref{fig:FPR} might be interpreted that naive inference with $K=15$ and $20$ could also control the FPRs properly, this was actually not the case. 
Under the null hypothesis, the $p$-value should be uniformly distributed between 0 and 1 (see, e.g., Section 3 in \citet{lehmann2006testing}). 
Figure \ref{fig:histogram} shows the distributions of (a) the selective $p$-values and (b) the naive $p$-values, where we see that the former are uniformly distributed, while the latter are not uniformly distributed.
Indeed, the Kolmogorov-Smirnov test for uniformity resulted in $p=0.314$ for selective $p$-values, but $p=10^{-6}$ for naive $p$-values. 

\subsubsection{FPRs of existing methods}
As mentioned in \S\ref{sec:setup_relatedworks}, there are two existing CP detection methods for multi-dimensional sequences.
In both methods, the asymptotic sampling distribution of the test statistic in the form of \eq{eq:test-statistic} is derived under certain assumptions.
Here, we see how these existing methods behave when the assumptions are violated. 

\begin{figure}[t!]
\centering
\vskip -.28in
\subfigure[$\ell_\infty$-aggregation]{
\includegraphics[width=.5\linewidth, viewport=0 0 360 360]{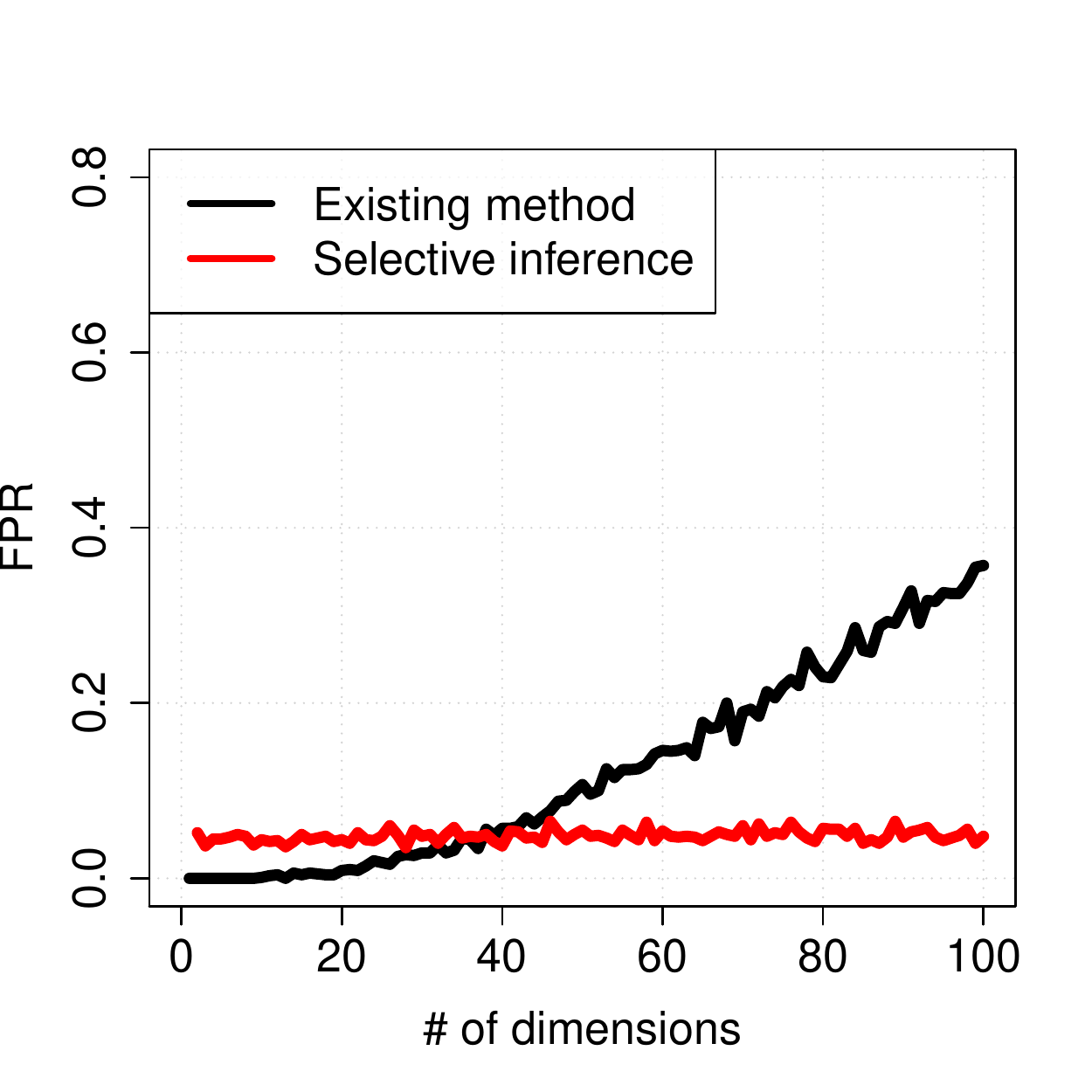}
\label{subfig:Jirak}
}
\hskip -.175in
\subfigure[kernel CP detection]{
\includegraphics[width=.5\linewidth, viewport=0 0 360 360]{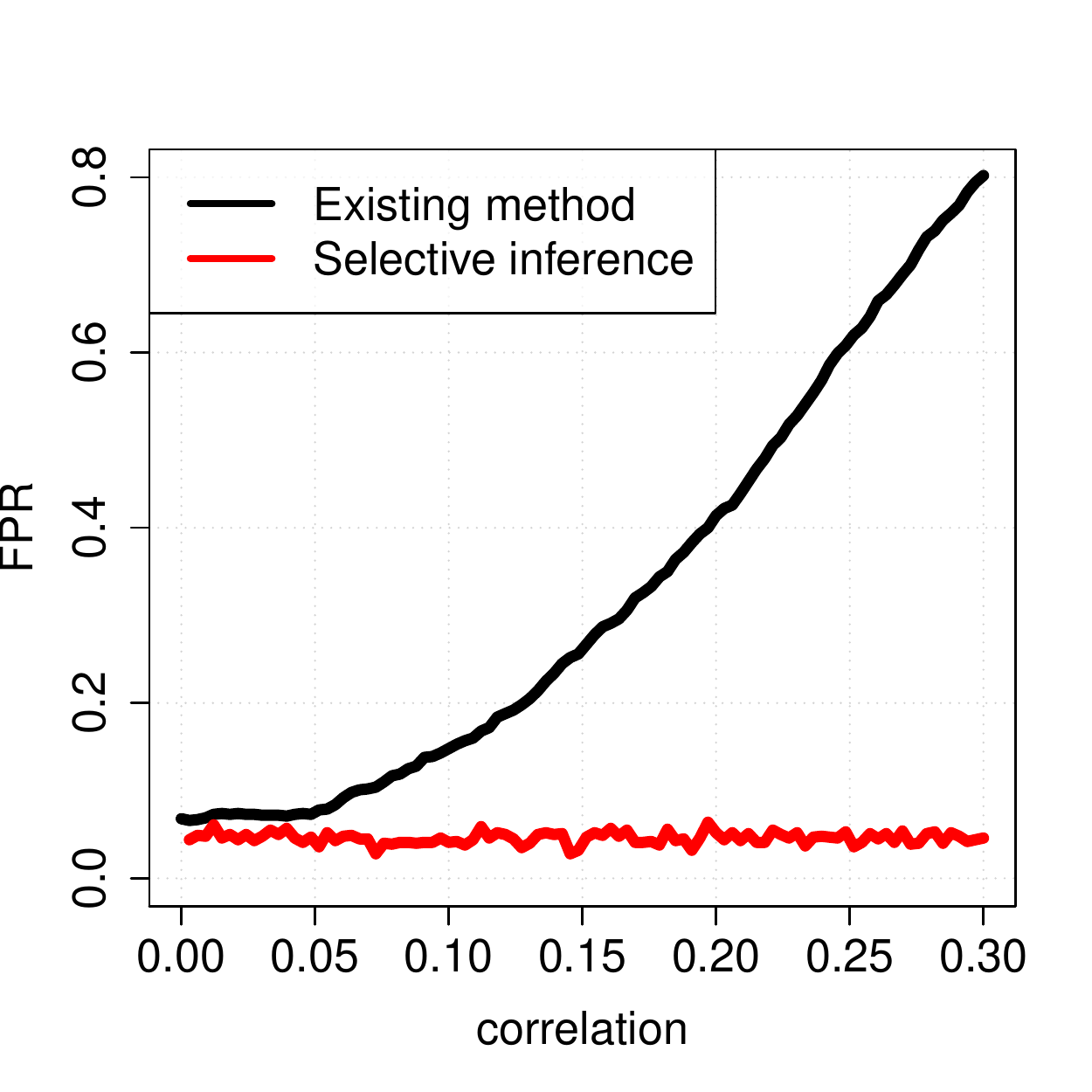}
\label{subfig:MMD}
}
\vskip -.1in
\caption{
 FPRs of two existing CP detection methods for a multi-dimensional sequence. 
 In (a), the FPRs of the method by \citet{Jir15} are plotted versus the number of dimensions. 
 In (b), the FPRs of the method by \citet{li2015m} are plotted versus the degree of correlation among different time points. 
 In both plots, the existing methods failed to control FPRs to the desired level when the underlying assumptions are violated. 
 On the other hand, the proposed selective inference framework successfully controlled the FPRs in all settings. 
}
\vskip -.2in
\end{figure}

First, to see the performances of the method proposed in \citet{Jir15}, we generated $Y$ from $\text{N}_{NT}({\rm vec}(O), I_T \times I_N)$ with $T=100$, and investigated the performances as $N$ varies from 1 to 100 (see \figurename~\ref{subfig:Jirak}).
We observe that the FPRs increase as $N$ becomes large when the assumption in \citet{Jir15} is violated. 
In contrast, the proposed selective inference (with double CUSUM aggregation) could appropriately control FPRs to the desired significant level of $0.05$ in the same setting (red solid line).

Next, to see the performances of the method proposed in \citet{li2015m}, we generated $Y$ from $\text{N}_{NT}({\rm vec}(O), \Xi \times I_N)$, where $\Xi=(\xi^{|i-j|})_{i,j\in[T]}$ with $N=10$ and $T=100$, and investigated the performances as $\xi$ varies from 0.0 to 0.3 (see \figurename~\ref{subfig:MMD}).
We observe that the FPRs increase as $\xi$ increases when the assumption in \citet{li2015m} is violated. 
Again, the proposed selective inference could appropriately control FPRs to the desired significant level of $0.05$ in the same setting.

\subsection{Application to CNV Detection}
\label{sec:real}
We applied the proposed selective inference framework for WRAG methods to a copy number variation (CNV) study on malignant lymphoma \citep{takeuchi2009potential}. 
In this study, CNVs of 46 patients diagnosed with diffuse large B-cell lymphoma (DLBCL) were investigated by an array comparative genomic hybridization (array CGH) technique \citep{hodgson2001genome}. 
The dataset that we analyze here is represented as a real-valued multi-dimensional sequence with $N=46$ and $T=2167$. 
Each dimension indicates a patient, while each time point indicates a local genomic region. 
It is well known that CNVs in DLBCL are heterogeneous because DLBCL has several subtypes\footnote{Identifying and characterizing the genetic properties of disease subtypes are crucially important for precision medicine.}. 
The goal of this medical study is to detect CNVs commonly observed in a subset of patients. 
Various one-dimensional CP detection methods have been used for analyzing array CGH data for a single patient \citep{wang2005method, tibshirani2008spatial, rapaport2008classification}.
However, there is no existing method for detecting common CPs by analyzing CNVs of multiple patients altogether, or for providing the statistical significance of the detected CNVs.

Due to space limitations, we only present the results for Chromosome 1, in which there are $T=177$ local genomic regions. 
We applied a WRAG method with a double CUSUM aggregation function to this dataset. 
For detecting multiple CPs, we used a local hypothesis testing framework described in \S\ref{sec:extension}, in which we set $h=5$ because this is the closest integer to $\log(T)$.
The covariance structure $\Sigma$ was set to be $I_N$ because each dimension in this multi-dimensional sequence was obtained from an individual patients. 
On the other hand, the covariance structure $\Xi$ was estimated from a different control dataset with $N=18$\footnote{CNV data in array CGH analysis is obtained by comparing the CNs between the patient and a healthy reference person. Therefore, a control dataset (without any CNVs) can be easily obtained by comparing the CNs between two healthy reference persons.}.
The parameter $\varphi$ in double CUSUM was set to be 0.5 as suggested in \citet{Cho16}. 

\begin{figure}[t!]
\centering
\vskip -.3in
\subfigure[
 Changes are observed in patients 3, 4, 21 and 40 at the 22nd genomic region with selective $p$-value 0.000.
 ]{
\includegraphics[scale=.275, viewport=0 0 720 432]{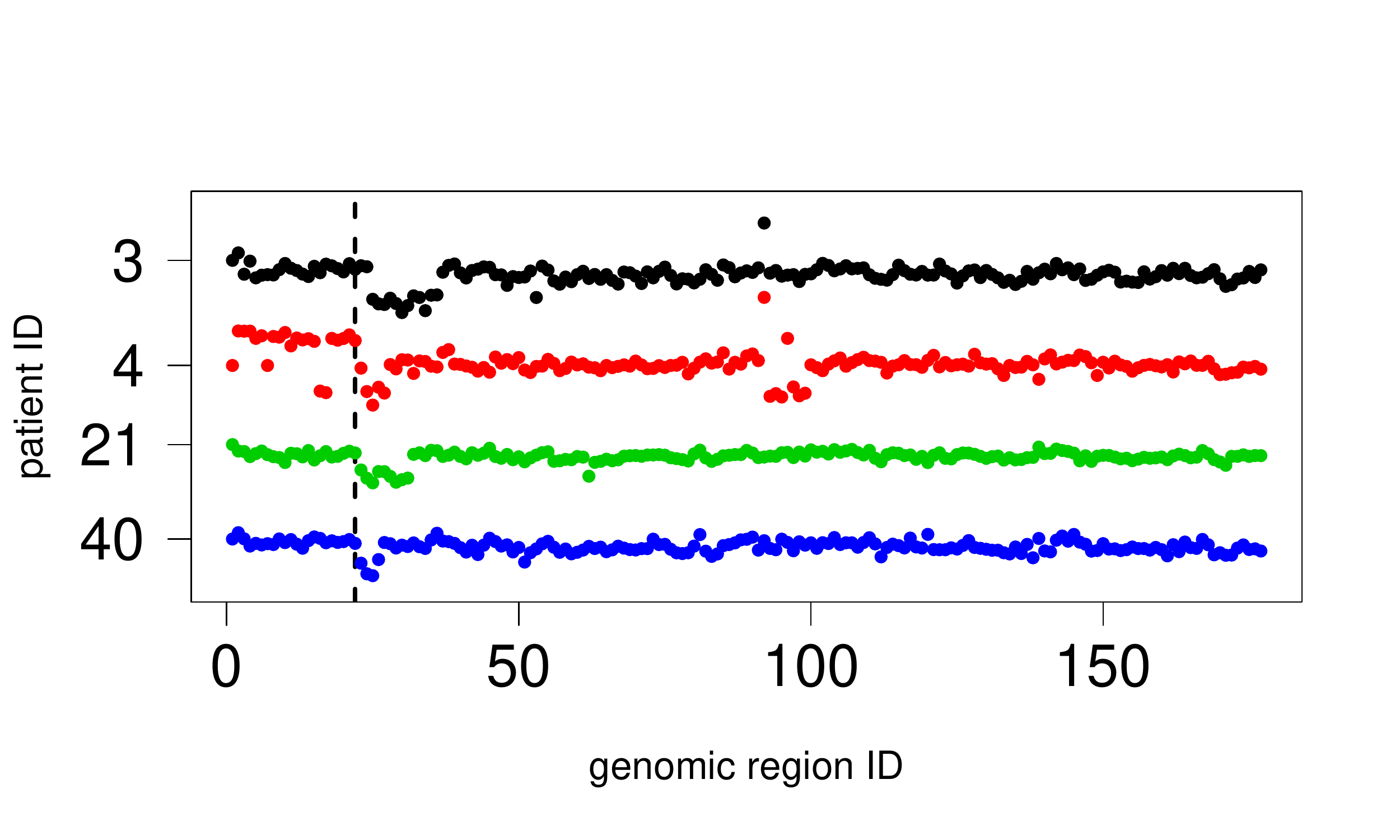}
\label{fig:negative-jump}
}
\vskip -.3in
\subfigure[
 Changes are observed in patients 2 and 21 at the 31st genomic region with selective $p$-value 0.001.
 ]{
\includegraphics[scale=.275, viewport=0 0 720 432]{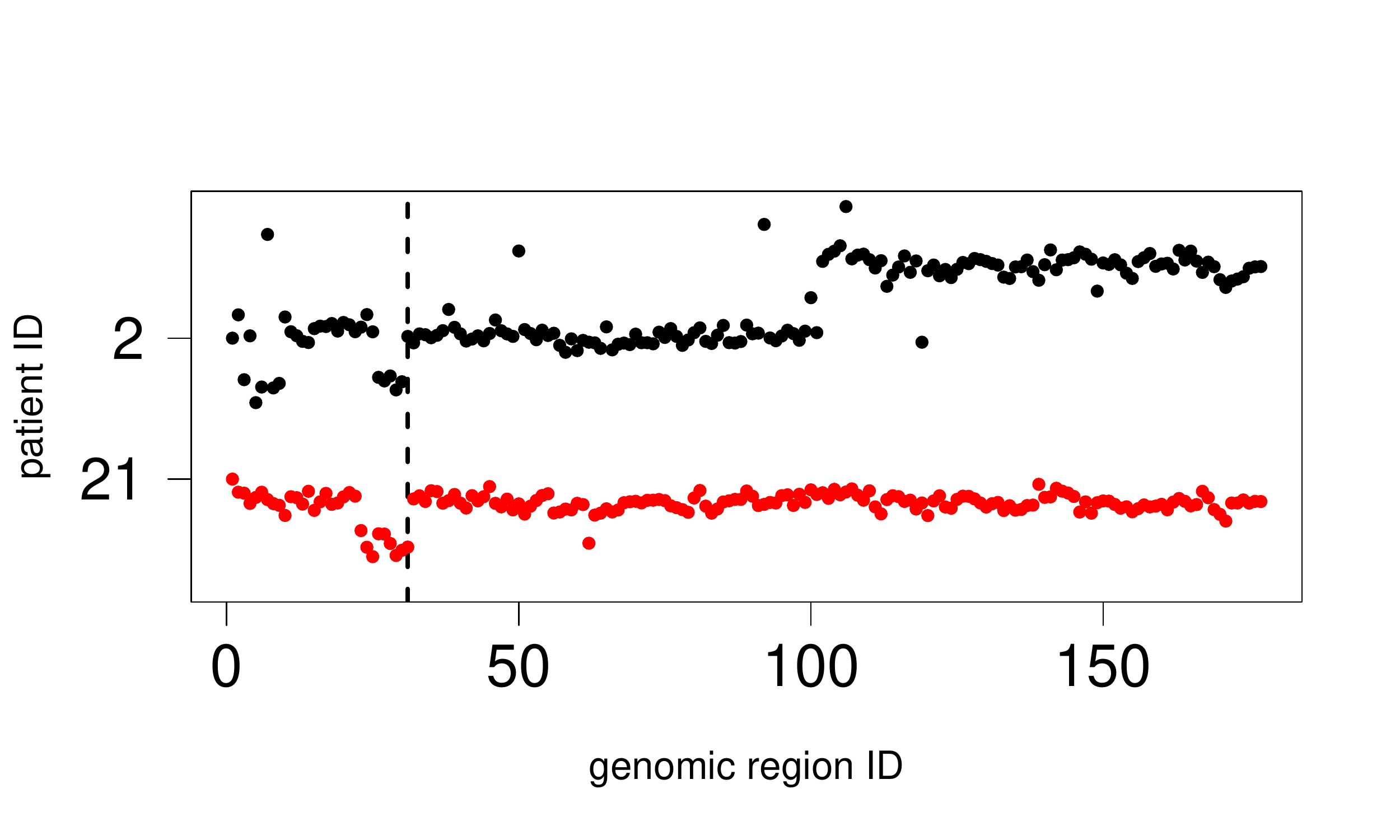}
\label{fig:positive-jump}
}
\vskip -.1in
\caption{
Examples of the detected CPs. 
} 
\vskip -.2in
\end{figure}

We detected 54 CPs and 11 of them are statistically significant in the sense that the selective $p$-value is less than 0.05.
Table \ref{tab:positive-jump} shows the list of detected CPs.
Two examples of the detected CPs are illustrated in Figure \ref{fig:positive-jump}.
Note that the numbers of the selected dimensions (patients) are different among the detected CPs, which is an advantage of the double CUSUM aggregation function. 
Our selective inference interpretation of CP detection problems allows us to properly correct the selection bias even if the selection procedure is fairly complicated as in double CUSUM aggregation. 

\begin{table}[t!]
\vskip -.15in
\caption{Detected copy number variations in Chromosome 1. The list of genomic region IDs for the array CGH analysis in \cite{takeuchi2009potential}, known genes in the genomic region, \# of selected  patients (i.e., $k$ in double CUSUM aggregation), and selective $p$-values of each detected CPs are shown.}
\label{tab:positive-jump}
\vskip -0.75in
\begin{center}
\begin{scriptsize}
\begin{sc}
\begin{tabular}{cl@{\hspace{-.2in}}c@{\hspace{-.1in}}c}
\toprule
\begin{tabular}{cc} genomic \\ region ID \end{tabular} & gene name & \begin{tabular}{cc} \# of selected \\ patients \end{tabular}& \begin{tabular}{cc} selective \\ $p$-value \end{tabular} \\
\midrule
7 &Q8N7E4 & 5 & 0.010 \\
15 &NM018125 & 5 & 0.028 \\
18 &CDA/KIF17 & 1 & 0.000 \\
22 &PAFAH2 & 4 & 0.000 \\
31 & EIF2C1 & 2 & 0.001 \\
36 & NA & 1 & 0.000 \\
106& \hskip-.1in \begin{tabular}{ll} SH2D2A/INSRR \\ /NTRK1 \end{tabular}& 13 & 0.038 \\
120 & C1orf9/TNFSF6 & 21 & 0.040 \\
151 & RPS6KC1 & 1 & 0.010 \\
162 & PSEN2 & 1 & 0.000 \\
165 &DISC1 & 23 & 0.044 \\
\bottomrule
\end{tabular}
\end{sc}
\end{scriptsize}
\end{center}
\vskip -0.2in
\end{table}

\bibliographystyle{icml2018}
\bibliography{ref}

\clearpage

\onecolumn
\appendix

\section{Example of Truncation Points}
\label{sec:example}
In this section, we evaluate truncation points $L$ and $U$ in Theorem \ref{thm2} for several aggregation functions described in \S \ref{sec:propose}, i.e., general WRAG, $\ell_\infty$-aggregation, $\ell_1$-aggregation and top $K$-aggregation functions.

\subsection{General WRAG}
\label{subsec:WRAG}
To derive the truncation points $L$ and $U$ in Theorem \ref{thm2}, we first show that $\theta_k(t)$ can be expressed as a bi-linear form of $Y$.
By the definition of multi-variate CUSUM score, we have $\bm{S}(t)=Y\bm{\eta}_t$, where
\begin{align*}
\bm{\eta}_t=\left\{\frac{T}{t(T-t)}\right\}^{1/2}\left(\frac{1}{t}\bm{1}_t^\top, -\frac{1}{T-t}\bm{1}_{T-t}^\top\right)^\top\in\mathbb{R}^T.
\end{align*}
Let $P_t=\text{diag}(\text{sgn}(S_i(t)))$ be an $N$-by-$N$ diagonal matrix whose diagonal elements are the sign of $\{S_i(t)\}_{i\in[N]}$ for each $t$.
Then, by the definition of $\rho_1(t),\ldots, \rho_N(t)$, there exists $N$-by-$N$ permutation matrix $G_t$ such that $\bm{\rho}(t)=G_t P_t \bm{S}(t)$.
Combining all the above, (\ref{eq:selected-test-statistic}) can be reduced to 
\begin{align*}
\theta_k(t)
=\bm{c}_k^\top\bm{\rho}(t)
=\bm{c}_k^\top G_t P_t Y\bm{\eta}_t,
\end{align*}
where $\bm{c}_k=(c_{j,1},\ldots, c_{k,N})^\top$.
Let $\bm{\delta}_{k,t}=P_t G_t^\top\bm{c}_k$.
Then, the selection event would be expressed as an affine constraint with respect to $\text{vec}(Y)$ (see, Section \ref{subsec:proof-thm2}).
Precisely, let  $D$ be a first order difference matrix, that is, 
\begin{align*}
D=
\left( \begin{array}{cccccc}
1 & -1 & 0 & \cdots & 0 & 0 \\
0 & 1 & -1 & \cdots & 0 & 0 \\
\vdots & \vdots & \vdots & \ddots & \vdots & \vdots \\
0 & 0 & 0 & \cdots & 1 & -1
\end{array} \right)
\in\mathbb{R}^{(N-1)\times N}.
\end{align*}
Then, the event 
\begin{align*}
\cE=\bigl\{(\hat{t},\hat{k})=(t,k)\bigr\}
\bigcap_{u\in[T-1]}\bigl\{(\hat{G}_u, \hat{P}_u)=(G_u, P_u)\bigr\}
\end{align*}
would be reduced to the intersection of affine constraints $\{A_{i,u}\text{vec}(Y)\leq \bm{0}\}~(i=1,2,3)$ for all $u\in[T-1]$, where
\begin{align*}
A_{1,u}
&=(\bm{\eta}_u\otimes \bm{\delta}_{l,u}- \bm{\eta}_t\otimes \bm{\delta}_{k,t})_{l\in[N-1]}^\top, \\
A_{2,u}
&=-\bm{\eta}_u^\top \otimes G_uP_u
\intertext{and}
A_{3,u}
&=-\bm{\eta}_u^\top\otimes DG_uP_u.
\end{align*}

To derive truncation points, let $\bm{\gamma}=(\Xi\bm{\eta}_t\otimes\Sigma\bm{\delta}_{k,t})/(\|\bm{\eta}_t\|_\Xi^2\|\bm{\delta}_{k,t}\|_\Sigma^2)$ and $\bm{z}=(I_{NT}-\bm{\gamma}(\bm{\eta}_t\otimes\bm{\delta}_{k,t})^\top)\text{vec}(Y)$.
Since $\theta_k(t)$ itself is non-negative, Lemma \ref{lem1} implies
\begin{align*}
L
&=\max\Biggl[
0,
\max_{l:(A\bm{\gamma})_l<0}\frac{b_v-(A\bm{z})_l}{(A\bm{\gamma})_l}
\Biggr] \\
&=\max\Biggl[
0,
\max_{u\in [T-1]}\max_{i=2,3,4}\max_{l:(A_{i,u}\bm{\gamma})_l<0}\frac{-(A_{i,u}\bm{z})_l}{(A_{i,u}\bm{\gamma})_l}
\Biggr] \\
&=\max\Bigl[0,\max_{u\in[T-1]}\max \bigl\{L_{1,u}, L_{2,u}, L_{3,u}\bigr\}+\theta_k(t)\Bigr],
\end{align*}
by simple calculations, where
\begin{subequations}
\begin{align}
\label{subeq:L1}
L_{1,u}
&=\max_{l:(A_{1,u}\bm{\gamma})_l<0}\frac{-(A_{1,u}\text{vec}(Y))_l}{(A_{1,u}\bm{\gamma})_l}, \\
\label{subeq:L2}
L_{2,u}
&=\max_{l:(A_{2, u}\bm{\gamma})_l<0}\frac{-(A_{2, u}\text{vec}(Y))_l}{(A_{2, u}\bm{\gamma})_l}
\intertext{and}
\label{subeq:L3}
L_{3,u}
&=\max_{l:(A_{3, u}\bm{\gamma})_l<0}\frac{-(A_{3, u}\text{vec}(Y))_l}{(A_{3, u}\bm{\gamma})_l}.
\end{align}
\end{subequations}
First, it hold that
\begin{align*}
(\bm{\eta}_u\otimes \bm{\delta}_{l,u}- \bm{\eta}_t\otimes \bm{\delta}_{k,t})^\top\text{vec}(Y)
=-\theta_k(t)+\theta_l(u)
~~~~~
\text{and}
~~~~~
(\bm{\eta}_u\otimes \bm{\delta}_{l,u}- \bm{\eta}_t\otimes \bm{\delta}_{k,t})^\top\bm{\gamma}
=\tau_{u,l}-1,
\end{align*}
where $\tau_{u,l}=(\bm{\eta}_u^\top\Xi\bm{\eta}_t\cdot \bm{\delta}_{l,u}^\top\Sigma \bm{\delta}_{k,t})/(\|\bm{\eta}_t\|_\Xi^2\|\bm{\delta}\|_\Sigma^2)$.
Thus we see that (\ref{subeq:L1}) implies
\begin{align*}
L_{1,u}
=\max_{l: \tau_{u,l}<1}\frac{\theta_k(t)-\theta_l(u)}{\tau_{u,l}-1}.
\end{align*}
In addition, we have
\begin{align*}
(-\bm{\eta}_u^\top \otimes G_uP_u)\text{vec}(Y)
=-\bm{\rho}(u)
~~~~~
\text{and}
~~~~~
(-\bm{\eta}_u^\top \otimes G_uP_u)\bm{\gamma}
=-\frac{\bm{\eta}_u^\top \Xi\bm{\eta}_t}{\|\bm{\eta}_t\|_\Xi^2\|\bm{\delta}_t\|_\Sigma^2} G_uP_u\Sigma\bm{\delta}_t
=:\bm{\omega}_u.
\end{align*}
Then, (\ref{subeq:L2}) can be reduced to
\begin{align*}
L_{2,u}
=\max_{l: \omega_{u,l}<0}\frac{\rho_l(u)}{\omega_{u,l}}.
\end{align*}
Finally,
\begin{align*}
(-\bm{\eta}_u^\top\otimes DG_uP_u)\text{vec}(Y)
=(-\rho_l(u)+\rho_{l+1}(u))_{l\in[N-1]}
~~~~~
\text{and}
~~~~~
(-\bm{\eta}_u^\top\otimes DG_uP_u)\bm{\gamma}
=D\bm{\omega}_u
\end{align*}
imply
\begin{align*}
L_{3,u}
=\max_{l: (D\bm{\omega}_u)_l<0}\frac{\rho_l(u)-\rho_{l+1}(u)}{(D\bm{\omega}_u)_l}
=\max_{l: \omega_{u,l}<\omega_{u,l+1}}\frac{\rho_l(u)-\rho_{l+1}(u)}{\omega_{u,l}-\omega_{u,l+1}}.
\end{align*}
Combining all the above, lower truncation point $L$ can be obtained by
\begin{align*}
L
&=\max \Biggl[0, 
\max_{u\in[T-1]}
\Biggl\{
\max_{l: \tau_{u,l}<1}\frac{\theta_k(t)-\theta_l(u)}{\tau_{u,l}-1},
\max_{l: \omega_{u,l}<0}\frac{\rho_l(u)}{\omega_{u,l}},
\max_{l: \omega_{u,l}<\omega_{u,l+1}}\frac{\rho_l(u)-\rho_{l+1}(u)}{\omega_{u,l}-\omega_{u,l+1}}
\Biggr\}
+\theta_k(t)
\Biggr].
\end{align*}
Similarly, by using the fact that the constraint $\{\theta_k(t)\geq 0\}$ does not affect to an upper truncation point, $U$ can be obtained by
\begin{align*}
U
&=\max_{u\in[T-1]}
\Biggl\{
\min_{l: \tau_{u,l}>1}\frac{\theta_k(t)-\theta_l(u)}{\tau_{u,l}-1},
\min_{l: \omega_{u,l}>0}\frac{\rho_i(u)}{\omega_{u,l}},
\min_{l: \omega_{u,l}>\omega_{u,l+1}}\frac{\rho_l(u)-\rho_{l+1}(u)}{\omega_{u,l}-\omega_{u,l+1}}
\Biggr\}
+\theta_k(t).
\end{align*} 

\subsection{$\ell_\infty$-aggregation}
\label{subsec:l_infty}
Recall that the aggregation function of $\ell_\infty$-aggregation score is expressed by
\begin{align*}
{\cal F}(\bm{S}(t))
=\max_{i\in[N]}|S_i(t)|.
\end{align*}
Let $\bm{\delta}_{i,t}=s_{i,t}\bm{e}_i$, where $s_{i,t}$ is the sign of $S_i(t)$ and $\bm{e}_i$ is an $N$-dimensional unit vector whose $i$-th element is one.
Then we see that $|S_i(t)|=\bm{\delta}_{i,t}^\top Y\bm{\eta}_t$, where $\bm{\eta}_t$ is an $N$-dimensional vector defined in Appendix \ref{subsec:WRAG}.
In $\ell_\infty$-aggregation, we consider the event $\{(\hat{t},\hat{i},\hat{s}_{i,t})=(t,i,s_{i,t})\}$ as a selection event, where $(\hat{t},\hat{i})$ is a maximizer of $|S_i(t)|$.
In this case, the constraint on the sign of $S_i(t)$ is equivalent to that on the non-negativity of test statistic.
Hence the event $\{(\hat{t},\hat{i},\hat{s}_{i,t})=(t,i,s_{i,t})\}$ can be expressed as 
\begin{align*}
\Bigl\{|S_i(t)|\geq |S_j(u)|,~{}^\forall u\in[T-1],~~\text{and}~{}^\forall j\in[N]\Bigr\}\cap\Bigl\{|S_i(t)|\geq 0\Bigr\}.
\end{align*}
Note that the former event in the above expression can be rewritten by
\begin{align*}
\{|S_i(t)|\geq |S_j(u)|,~{}^\forall u\in[T-1],~~\text{and}~{}^\forall j\in[N]\}
&=\bigcap_{u\in[T-1]}\bigcap_{j\in[N]}\Bigl\{|S_i(t)|\geq |S_j(u)|\Bigr\} \\
&=\bigcap_{u\in[T-1]}\bigcap_{j\in[N]}\Bigl\{-s_{i,t}S_i(t)\leq S_j(u)\leq s_{i,t}S_i(t)\Bigr\} \\
&=\bigcap_{u\in[T-1]}\bigcap_{j\in[N]}\Bigl\{-\bm{\delta}_{i,t}^\top Y\bm{\eta}_t\leq \bm{e}_j^\top Y\bm{\eta}_u\leq \bm{\delta}_{i,t}^\top Y\bm{\eta}_t\Bigr\} \\
&=\bigcap_{u\in[T-1]}\bigcap_{j\in[N]}
\left\{
\begin{array}{c}
-\bm{\delta}_{i,t}^\top Y\bm{\eta}_t-\bm{e}_j^\top Y\bm{\eta}_u\leq 0, \\
-\bm{\delta}_{i,t}^\top Y\bm{\eta}_t+\bm{e}_j^\top Y\bm{\eta}_u\leq 0
\end{array}
\right\}.
\end{align*}
To apply Lemma \ref{lem1}, define
\begin{align*}
\bm{\gamma}
=\frac{1}{\|\bm{\eta}_t\|_\Xi^2\|\bm{\delta}_{i,t}\|_\Sigma^2}(\Xi\bm{\eta}_t\otimes \Sigma\bm{\delta}_{i,t})
=\frac{s_{i,t}}{\|\bm{\eta}_t\|_\Xi^2\Sigma_{ii}}(\Xi\bm{\eta}_t\otimes \Sigma\bm{e}_i).
\end{align*}
Then, we can see that
\begin{align*}
-\bm{\delta}_{i,t}^\top Y\bm{\eta}_t-\bm{e}_j^\top Y\bm{\eta}_u\leq 0
\Leftrightarrow (-\bm{\eta}_t\otimes \bm{\delta}_{i,t}-\bm{\eta}_u\otimes \bm{e}_j)^\top\text{vec}(Y)\leq 0.
\end{align*}
Therefore, we have
\begin{align*}
(-\bm{\eta}_t\otimes \bm{\delta}_{i,t}-\bm{\eta}_u\otimes \bm{e}_j)^\top\text{vec}(Y)
=-|S_i(t)|-S_j(u)
~~~~~
\text{and}
~~~~~
(-\bm{\eta}_t\otimes \bm{\delta}_{i,t}-\bm{\eta}_u\otimes \bm{e}_j)^\top\bm{\gamma}
=-1-\tau_{u,j},
\end{align*}
where
\begin{align*}
\tau_{u,j}
=\frac{s_{i,t}}{\|\bm{\eta}_t\|_\Xi^2\Sigma_{ii}}\bm{\eta}_u^\top\Xi\bm{\eta}_t\cdot\bm{e}_j^\top\Sigma\bm{e}_i
=\frac{s_{i,t}\Sigma_{ji}}{\|\bm{\eta}_t\|_\Xi^2\Sigma_{ii}}\bm{\eta}_u^\top\Xi\bm{\eta}_t.
\end{align*}
Similarly, 
\begin{align*}
-\bm{\delta}_{i,t}^\top Y\bm{\eta}_t+\bm{e}_j^\top Y\bm{\eta}_u\leq 0
\Leftrightarrow (-\bm{\eta}_t\otimes \bm{\delta}_{i,t}+\bm{\eta}_u\otimes \bm{e}_j)^\top\text{vec}(Y)\leq 0,
\end{align*}
and thus, we have
\begin{align*}
(-\bm{\eta}_t\otimes \bm{\delta}_{i,t}+\bm{\eta}_u\otimes \bm{e}_j)^\top\text{vec}(Y)
=-|S_i(t)|+S_j(u)
~~~~~
\text{and}
~~~~~
(-\bm{\eta}_t\otimes \bm{\delta}_{i,t}+\bm{\eta}_u\otimes \bm{e}_j)^\top\bm{\gamma}
=-1+\tau_{u,j}.
\end{align*}
Hence, we obtain, from Lemma \ref{lem1}, that
\begin{align*}
L
&=\max\Biggl[
0,
\max_{u\in[T-1]}
\Biggl\{
\max_{i:\tau_{u,j}>-1}\frac{|S_i(t)|+S_j(u)}{-1-\tau_{u,j}}, 
\max_{i:\tau_{u,j}<1}\frac{|S_i(t)|-S_j(u)}{-1+\tau_{u,j}}
\Biggr\}
+|S_i(t)|
\Biggr] \\
&=\max\Biggl[ 0,
\max_{u\in[T-1]}
\Biggl\{
\max_{j:\tau_{u,j}>-1}\frac{\tau_{u,j}|S_i(t)|-S_j(u)}{\tau_{u,j}+1},
\max_{j:\tau_{u,j}<1}\frac{\tau_{u,j}|S_i(t)|+S_j(u)}{\tau_{u,j}-1}
\Biggr\}
\Biggr]
\intertext{and, since $|S_i(t)|=\bm{\delta}_{i,t}^\top Y\bm{\eta}_t$ itself is non-negative and this does not affect to upper truncation point, we obtain}
U
&=\min_{u\in[T-1]}
\Biggl\{
\min_{j:\tau_{u,j}<-1}\frac{\tau_{u,j}|S_i(t)|-S_j(u)}{\tau_{u,j}+1}, 
\min_{j:\tau_{u,j}>1}\frac{\tau_{u,j}|S_i(t)|+S_j(u)}{\tau_{u,j}-1}
\Biggr\}.
\end{align*}

\subsection{$\ell_1$-aggregation}
\label{subsec:l_1}
Recall that the aggregation function of $\ell_1$-aggregation score is expressed by
\begin{align*}
{\cal F}(\bm{S}(t))
=\sum_{i\in[N]}|S_i(t)|.
\end{align*}
Let $\bm{1}_N$ be an $N$-dimensional one-vector, and let $\bm{\delta}_{t}=\bm{1}_N^\top P_t$, where $P_t$ is an $N$-by-$N$  diagonal matrix defined in Appendix \ref{subsec:WRAG}.
Then we see that $\sum_{i\in[N]}|S_i(t)|=\bm{\delta}_t^\top Y\bm{\eta}_t$.
In the following, we consider the event $\bigcap_{u\in[T-1]}\{(\hat{t}, \hat{P}_u)=(t, P_u)\}$ as a selection event.
Similar to Appendix \ref{subsec:WRAG} and \ref{subsec:l_infty}, the selection event can be expressed as
\begin{align*}
\bigcap_{u\in[T-1]}\Biggl[
\Biggl\{\sum_{i\in[N]}|S_i(t)|\geq \sum_{i\in[N]}|S_i(u)|\Biggr\}
\cap
\Bigl\{|S_i(u)|\geq 0,
~
{}^\forall i\in[N] \Bigr\}
\Biggr].
\end{align*}
Here, the last constraint corresponds to the conditions on $P_u$.

First, it hold that
\begin{align*}
\Biggl\{\sum_{i\in[N]}|S_i(t)|\geq \sum_{i\in[N]}|S_i(u)|\Biggr\}
&=\Biggl\{\sum_{i\in[N]}|S_i(t)|\geq \sum_{i\in[N]}|S_i(u)|\Biggr\} \\
&=\Bigl\{-\bm{\delta}_t^\top Y\bm{\eta}_t+ \bm{\delta}_u^\top Y\bm{\eta}_u\leq 0\Bigr\},
\intertext{and}
\Bigl\{|S_i(u)|\geq 0,~{}^\forall i\in[N] \Bigr\} 
&=\Bigl\{-P_u Y\bm{\eta}_u\leq \bm{0} \Bigr\}.
\end{align*}
To apply Lemma \ref{lem1}, define
$\bm{\gamma}=(\Xi\bm{\eta}_t\otimes \Sigma\bm{\delta}_t)/(\|\bm{\eta}_t\|_\Xi^2\|\bm{\delta}_t\|_\Sigma^2)$.
Then, we see that
\begin{align*}
-\bm{\delta}_t^\top Y\bm{\eta}_t+ \bm{\delta}_u^\top Y\bm{\eta}_u\leq 0
\Leftrightarrow (-\bm{\eta}_t\otimes \bm{\delta}_t+\bm{\eta}_u\otimes \bm{\delta}_u)^\top\text{vec}(Y)\leq 0.
\end{align*}
Therefore, we have
\begin{align*}
(-\bm{\eta}_t\otimes \bm{\delta}_t+\bm{\eta}_u\otimes \bm{\delta}_u)^\top\text{vec}(Y)
&=-\sum_{i=1}^N|S_i(t)|+\sum_{i=1}^N|S_i(u)| \\
&=-\sum_{i=1}^N(|S_i(t)|-|S_i(u)|)
\intertext{and}
(-\bm{\eta}_t\otimes \bm{\delta}_t+\bm{\eta}_u\otimes \bm{\delta}_u)^\top\bm{\gamma}
&=-1+\tau_u,
\end{align*}
where
$\tau_{u}=\bm{\eta}_u^\top\Xi\bm{\eta}_t\cdot\bm{\delta}_u^\top\Sigma\bm{\delta}_t/(\|\bm{\eta}_t\|_\Xi^2\|\bm{\delta}_t\|_\Sigma^2)$.
Similarly, 
\begin{align*}
-P_u Y\bm{\eta}_u\leq \bm{0}
\Leftrightarrow (-\bm{\eta}_u\otimes P_u)^\top\text{vec}(Y)\leq \bm{0},
\end{align*}
and hence we have
\begin{align*}
(-\bm{\eta}_u\otimes P_u)^\top\text{vec}(Y)
&=-P_t\bm{S}(u)
=-(|S_l(u)|)_{l\in[N]}
\intertext{and}
(-\bm{\eta}_u\otimes P_u)^\top\bm{\gamma}
&=-\frac{\bm{\eta}_u^\top\Xi\bm{\eta}_t}{\|\bm{\eta}_t\|_\Xi^2\|\bm{\delta}_t\|_\Sigma^2}P_u\Sigma\bm{\delta}_t
=:\bm{\omega}_u.
\end{align*}
Finally, since $\sum_{i\in[N]}|S_i(t)|=\bm{\delta}_t^\top Y\bm{\eta}_t$ is non-negative and thus lower truncation point is non-negative, we obtain, from Lemma \ref{lem1}, that
\begin{align*}
L
&=\max\Biggl[0,
\max
\Biggl\{
\max_{u:\tau_u>-1}\frac{\sum_{i\in[N]}(|S_i(t)|-|S_i(u)|)}{-1+\tau_u},
\max_{u\in[T-1]}\max_{l:\omega_{u,l}<0}\frac{|S_l(u)|}{\omega_{u,l}}
\Biggr\}
+\sum_{i\in[N]}|S_i(t)|
\Biggr] \\
&=\max\Biggl[0,
\sum_{i\in[N]} 
\max
\left\{
\max_{u:\tau_u>-1}\frac{\tau_u|S_i(t)|-|S_i(u)|}{\tau_u-1},
\max_{u\in[T-1]}\max_{l:\omega_{u,l}<0}
\frac{|S_l(u)|}{N\omega_{u,l}}+|S_i(t)|\right\}
\Biggr]
\intertext{and}
U
&= \sum_{i\in[N]}
\min
\left\{
\min_{u:\tau_u<-1}\frac{\tau_u|S_i(t)|-|S_i(u)|}{\tau_u-1},
\min_{u\in[T-1]}
\min_{l:\omega_{u,l}>0}
\frac{|S_l(u)|}{N\omega_{u,l}}+|S_i(t)|
\right\}.
\end{align*}

\subsection{Top $K$-aggregation}
To derive truncation points of top $K$-aggregation, we first introduce following lemma.
\begin{lemm}
\label{lem4}
Let $\bm{x}=(x_1,\ldots, x_N)^\top \in\mathbb{R}_+^N$ and $I=\{i_1,\ldots, i_K\}\subset [N]$ with $|I|=K$.
Define
\begin{align*}
A&=
\Bigl\{
\bm{x}\in\mathbb{R}_+^N \mid
x_{i_1}+\cdots+x_{i_K}\geq x_{j_1}+\cdots+x_{j_K},
{}^\forall (j_1,\ldots,j_K)\subset[N]
\Bigr\}
\intertext{and}
B&=\Bigl\{\bm{x}\in\mathbb{R}_+^N \mid x_i\geq x_j,~{}^\forall (i,j)\in I\times I^c\Bigr\}.
\end{align*}
Then, $A=B$.
\end{lemm}

Recall that the aggregation function of top $K$-aggregated score is expressed by
\begin{align*}
{\cal F}(\bm{S}(t))
=\max_{I\subset[N]:|I|=K}\sum_{i\in I}|S_i(t)|.
\end{align*}
Let $\bm{\delta}_{I,t}=\bm{e}_I^\top P_{I, t}$, where $\bm{e}_I=(I\{i\in I\})_{i\in[N]}$ is an $N$-dimensional vector corresponding to $I\subset [N]$.
For simplicity of the notation, we require the signs of $S_i(t)$ for $i \in[N]$ although we can also justify the following statement for $i\in I$.
First, we see that $\sum_{i\in I}|S_i(t)|=\bm{\delta}_{I,t}^\top Y\bm{\eta}_t$.
In the following, we consider the event $\{(\hat{t},\hat{I},\hat{P}_{I,t})=(t,I,P_{I,t})\}$ as a selection event, where $\hat{I}$ is a maximizer of the above maximization problem.
Then, the event can be expressed by
\begin{align*}
\Biggl\{\sum_{i\in I}|S_i(t)|\geq \sum_{i\in J}|S_i(u)|,~{}^\forall u\in[T-1]~\text{and}~{}^\forall J\in[N]\Biggr\}
\cap\Bigl\{|S_i(t)|\geq 0,~{}^\forall i\in I\Bigr\}.
\end{align*}
First, from Lemma \ref{lem4} and the same argument as in Section \ref{subsec:l_infty}, we see that
\begin{align*}
\Biggl\{\sum_{i\in I}|S_i(t)|\geq \sum_{i\in J}|S_i(u)|,~{}^\forall u\in[T-1]~\text{and}~{}^\forall J\in[N]\Biggr\}
&=\bigcap_{u\in[T-1]}\Biggl\{\sum_{i\in I}|S_i(t)|\geq \sum_{i\in J}|S_i(u)|,~{}^\forall J\in[N]\Biggr\} \\
&=\bigcap_{u\in[T-1]}\Bigl\{|S_i(t)|\geq |S_j(u)|,~(i,j)\in I\times I^c\Bigr\} \\
&=\bigcap_{u\in[T-1]}\bigcap_{(i,j)\in I\times I^c}\Bigl\{|S_i(t)|\geq |S_j(u)|\Bigr\} \\
&=\bigcap_{u\in[T-1]}\bigcap_{(i,j)\in I\times I^c}
\left\{
\begin{array}{c}
-s_{i,t}\bm{e}_i^\top Y\bm{\eta}_t-\bm{e}_j^\top Y\bm{\eta}_u\leq 0, \\
-s_{i,t}\bm{e}_i^\top Y\bm{\eta}_t+\bm{e}_j^\top Y\bm{\eta}_u\leq 0
\end{array}
\right\}.
\end{align*}
In addition, we have $\{|S_i(t)|\geq 0\}=\{-\bm{\delta}_{I,t}^\top Y\bm{\eta}_t\leq 0\}$.
To apply Lemma \ref{lem1}, define
$\bm{\gamma}=(\Xi\bm{\eta}_t\otimes \Sigma\bm{\delta}_{I,t})/(\|\bm{\eta}_t\|_\Xi^2\|\bm{\delta}_{I,t}\|_\Sigma^2)$.
Then, we can see that
\begin{align*}
-s_{i,t}\bm{e}_i^\top Y\bm{\eta}_t-\bm{e}_j^\top Y\bm{\eta}_u\leq 0 
\Leftrightarrow (-s_{i,t}\bm{\eta}_t\otimes \bm{e}_i-\bm{\eta}_u\otimes \bm{e}_j)^\top\text{vec}(Y)\leq 0,
\end{align*}
and we have
\begin{align*}
(-s_{i,t}\bm{\eta}_t\otimes \bm{e}_i-\bm{\eta}_u\otimes \bm{e}_j)^\top\text{vec}(Y)
=-|S_i(t)|-S_j(u)
~~~~~
\text{and}
~~~~~
(-s_{i,t}\bm{\eta}_t\otimes \bm{e}_i-\bm{\eta}_u\otimes \bm{e}_j)^\top\bm{\gamma}
=-\tau_{i,t}-\tau_{j,u},
\end{align*}
where
$\tau_{j,u}=\bm{\eta}_u^\top\Xi\bm{\eta}_t\cdot\bm{e}_j^\top\Sigma\bm{\delta}_{I,t}/(\|\bm{\eta}_t\|_\Xi^2\|\bm{\delta}_{I,t}\|_\Sigma^2)$, for each $j\in [N]$ and $u\in[T-1]$.
Similarly, 
\begin{align*}
-s_{i,t}\bm{e}_i^\top Y\bm{\eta}_t+\bm{e}_j^\top Y\bm{\eta}_u\leq 0 
\Leftrightarrow (-s_{i,t}\bm{\eta}_t\otimes \bm{e}_i+\bm{\eta}_u\otimes \bm{e}_j)^\top\text{vec}(Y)\leq 0,
\end{align*}
and thus, we have
\begin{align*}
(-s_{i,t}\bm{\eta}_t\otimes \bm{e}_i+\bm{\eta}_u\otimes \bm{e}_j)^\top\text{vec}(Y)
=-|S_i(t)|+S_j(u)
~~~~~
\text{and}
~~~~~
(-s_{i,t}\bm{\eta}_t\otimes \bm{e}_i+\bm{\eta}_u\otimes \bm{e}_j)^\top\bm{\gamma}
=-\tau_{i,t}+\tau_{j,u}.
\end{align*}
Moreover, it hold that
\begin{align*}
\Bigl\{|S_i(t)|\geq 0,~{}^\forall i\in I\Bigr\}
&=\bigcap_{i\in I}\Bigl\{-s_{i,t}\bm{e}_i^\top Y\bm{\eta}_t\leq 0 \Bigr\} \\
&=\bigcap_{i\in I}\Bigl\{(-s_{i,t}\bm{\eta}_t\otimes \bm{e}_i)^\top\text{vec}(Y)\leq 0 \Bigr\},
\end{align*}
and thus we have
\begin{align*}
(-s_{i,t}\bm{\eta}_t\otimes \bm{e}_i)^\top\text{vec}(Y)
=-|S_i(t)|
~~~~~
\text{and}
~~~~~
(-s_{i,t}\bm{\eta}_t\otimes \bm{e}_i)^\top\bm{\gamma}
=-\tau_{i,t}.
\end{align*}
Therefore, from Lemma \ref{lem1}, we obtain
\begin{align*}
L
&=\max \Biggl[
0, 
\max_{u\in[T-1]}
\Biggl\{ \max_{(i,j)\in I\times I^c:\tau_{j,u}>-\tau_{i,t}}\frac{|S_i(t)|+S_j(u)}{-\tau_{i,t}-\tau_{j,u}},
\max_{(i,j)\in I\times I^c:\tau_{j,u}<\tau_{i,t}}\frac{|S_i(t)|-S_j(u)}{-\tau_{i,t}+\tau_{j,u}}
\Biggr\}
+\sum_{i\in I}|S_i(t)|, \\
&\hspace{1.5cm} \max_{i\in I: \tau_{i,t}> 0}\frac{|S_i(t)|}{-\tau_{i,t}}
+\sum_{i\in I}|S_i(t)|,
\Biggr]
\intertext{and}
U
&=\min\Biggl[
\min_{u\in[T-1]}\Biggl\{
\min_{(i,j)\in I\times I^c:\tau_{j,u}<-\tau_{i,t}}\frac{|S_i(t)|+S_j(u)}{-\tau_{i,t}-\tau_{j,u}},
\min_{(i,j)\in I\times I^c:\tau_{j,u}>\tau_{i,t}}\frac{|S_i(t)|-S_j(u)}{-\tau_{i,t}+\tau_{j,u}}
\Biggr\}, \\
&\hspace{1cm}  \min_{i\in I: \tau_{i,t}><0} \frac{|S_i(t)|}{-\tau_{i,t}}
\Biggr]
+\sum_{i\in I}|S_i(t)|.
\end{align*}

\section{Proofs of Main Result}
\label{sec:proof}
Here, we provide proofs of the main result in the paper.
We first show the proof of Lemma \ref{lem1}, and those of Theorem \ref{thm2} and \ref{thm3}.
\subsection{Proof of Lemma \ref{lem1}}
\label{subsec:proof-lem1}
The proof is similar to that in \citet{Lee16}.
Because $\text{vec}(Y)=\bm{z}+\bm{\gamma}(\bm{\eta}\otimes\bm{\delta})^\top\text{vec}(Y)$, we see that
\begin{align*}
A\text{vec}(Y)\leq \bm{b}
\Leftrightarrow
A\bm{\gamma}(\bm{\eta}\otimes\bm{\delta})^\top\text{vec}(Y)\leq \bm{b}-A\bm{z}.
\end{align*}
Hence, we have
\begin{align*}
&(\bm{\eta}\otimes\bm{\delta})^\top\text{vec}(Y) \geq \frac{(\bm{b}-A\bm{z})_j}{(A\bm{\gamma})_j}
~~~\text{if}~~~(A\bm{\gamma})_j<0, \\
&(\bm{\eta}\otimes\bm{\delta})^\top\text{vec}(Y) \leq \frac{(\bm{b}-A\bm{z})_j}{(A\bm{\gamma})_j}
~~~\text{if}~~~(A\bm{\gamma})_j>0
\intertext{and}
&0\leq (\bm{b}-A\bm{z})_j
~~~\text{if}~~~
(A\bm{\gamma})_j=0.
\end{align*}
From the above inequalities, we can conclude that $(\bm{\eta}\otimes\bm{\delta})^\top\text{vec}(Y)$ can be bounded below (resp. above) by $L(\bm{z})$ (resp. $U(\bm{z})$).
In addition, $(\bm{\eta}\otimes\bm{\delta})^\top\text{vec}(Y)=\bm{\delta}^\top Y\bm{\eta}$ yields last equality.
Moreover, since $\bm{z}$ is independent of $\bm{\delta}^\top Y\bm{\eta}$ under the normality of $Y$, that is, ${\rm Cov}(\bm{z}, \bm{\delta}^\top Y\bm{\eta})=0$, $(L(\bm{z}),U(\bm{z}),N(\bm{z}))$ is also independent of $\bm{\delta}^\top Y\bm{\eta}$.
 
\subsection{Proof of Theorem \ref{thm2}}
\label{subsec:proof-thm2}
Note that under the null hypothesis, $\hat{\theta}_k(t)$ is a random variable with mean zero, conditional on the selection event
\begin{align*}
\cE=\bigl\{(\hat{t},\hat{k})=(t,k)\bigr\}
\bigcap_{u\in[T-1]}\bigl\{(\hat{G}_u, \hat{P}_u)=(G_u, P_u)\bigr\}.
\end{align*}
To apply Lemma \ref{lem1}, we need to show that
\begin{itemize}
\item test statistic $\hat{\theta}_k(t)$ can be rewritten as a bi-linear form with respect to some two vectors, and
\item selection event $\cE$ can be reduced to an affine constraint with respect to $\text{vec}(Y)$.
\end{itemize}
If the above two claims are true, we can say that
\begin{align*}
[\hat{\theta}_k(t)\mid \cE, \bm{z}=\bm{z}_0] 
\sim \text{TN}(0, v^2, L(\bm{z}_0), U(\bm{z}_0)),
\end{align*}
where $\text{TN}(\mu, \sigma^2, l, u)$ denotes a normal distribution $\text{N}(\mu,\sigma^2)$ restricted to the interval $[l,u]$ and $v^2$ is a variance of $\theta_k(t)$ conditional on the selection event.
Moreover, by applying the probability integral transformation and then integrating out $\bm{z}_0$, the theorem hold.
See Theorem 5.2 in \citet{Lee16} for more details.

As shown in Appendix \ref{subsec:WRAG}, the former claim is true.
We now show the latter one.
Let $\bm{\delta}_{k,t}=P_t G_t^\top\bm{c}_k$ and $\bm{\eta}_t$ be $T$-dimensional vector and $N$-dimensional vector defined in Section \ref{subsec:WRAG}.
Since $\hat{t}$ and $\hat{k}$ are defined as the maximizer of (\ref{eq:selected-test-statistic}), the event $\cE$ is equivalent to
\begin{subequations}
\begin{align}
\label{subeq;SE1}
&\hat{\theta}_k(t)\geq \hat{\theta}_l(u),~~~{}^\forall l\in [N-1], \\
\label{subeq;SE2}
&\bm{\rho}(u)\geq \bm{0}
\intertext{and}
\label{subeq;SE3}
& \rho_1(u)\geq \cdots \geq \rho_N(u).
\end{align}
\end{subequations}
for all $u\in[T-1]$.
Intuitively, (\ref{subeq;SE1}), (\ref{subeq;SE2}) and (\ref{subeq;SE3}), respectively represents ``the test statistics is the maximizer of a optimization problem", ``signs of multi-variate CUSUM statistic" and ``the information about the permutation of the multi-variate CUSUM statistic".
First, (\ref{subeq;SE1}) implies
\begin{align*}
\bm{\delta}_{k,t}^\top Y\bm{\eta}_t\geq \bm{\delta}_{l,u}^\top Y\bm{\eta}_u
&\Leftrightarrow (\bm{\eta}_u\otimes \bm{\delta}_{l,u}- \bm{\eta}_t\otimes \bm{\delta}_{k,t})^\top\text{vec}(Y)\leq 0 \\
&\Leftrightarrow \bm{a}_{l, u}^\top\text{vec}(Y)\leq 0,
\end{align*}
for all $l\in[N-1]$, where $\bm{a}_{l, u}=\bm{\eta}_u\otimes \bm{\delta}_{l,u}- \bm{\eta}_t\otimes \bm{\delta}_{k,t}$.
Thus, by letting $A_{1,u}=(\bm{a}_{1, u}, \ldots, \bm{a}_{N-1, u})^\top$, we have
\begin{align*}
\bm{\delta}_{k,t}^\top Y\bm{\eta}_t\geq \bm{\delta}_{l,u}^\top Y\bm{\eta}_u,~~~{}^\forall l\in [N-1]
\Leftrightarrow  A_{1,u}\text{vec}(Y)\leq \bm{0}.
\end{align*}
Moreover, (\ref{subeq;SE2}) implies
\begin{align*}
G_uP_u Y\bm{\eta}_u\geq \bm{0}
&\Leftrightarrow -(\bm{\eta}_u\otimes P_uG_u^\top)^\top\text{vec}(Y)\leq \bm{0} \\
&\Leftrightarrow A_{2,u}\text{vec}(Y)\leq \bm{0}
\end{align*}
for each $u\in [T-1]$, where $A_{2,u}=-(\bm{\eta}_u\otimes P_uG_u^\top)^\top$.
By letting $D$ be a first order difference matrix, (\ref{subeq;SE3}) implies
\begin{align*}
D\bm{\rho}(u)\geq \bm{0}
&\Leftrightarrow -DG_u P_u Y\bm{\eta}_u\leq \bm{0} \\
&\Leftrightarrow -(\bm{\eta}_u\otimes P_uG_u^\top D^\top)^\top\text{vec}(Y)\leq \bm{0} \\
&\Leftrightarrow A_{3,u}\text{vec}(Y)\leq \bm{0}
\end{align*}
for each $u\in [T-1]$, where $A_{3,u}=-(\bm{\eta}_u\otimes P_uG_u^\top D^\top)^\top$.
Combining all the above, we have
\begin{align*}
\bigcap_{u\in[T-1]} 
\Bigl\{
A_{1, u}\text{vec}(Y)\leq \bm{0},
A_{2, u}\text{vec}(Y)\leq \bm{0},
A_{3, u}\text{vec}(Y)\leq \bm{0}
\Bigr\}
\end{align*}
which is an affine constraint.

Finally, by the definition of aggregated score, we have
\begin{align*}
\hat{\theta}_k(t)
=\bm{\delta}_{k,t}^\top Y\bm{\eta}_t
=(\bm{\eta}_t\otimes \bm{\delta}_{k,t})^\top\text{vec}(Y)
\geq 0, 
\end{align*}
and this states that the lower truncation point in (\ref{eq:truncation_points}) is always non-negative.

\subsection{Proof of Theorem \ref{thm3}}
\label{subsec:proof-thm3}
To state the proof, we require following lemmas.
\begin{lemm}
\label{lem5}
Let $x,y$ and $\sigma~(>0)$ be known constants and define $f(\mu)=\Phi((x-\mu)/\sigma)/\Phi((y-\mu)/\sigma)$, where $\Phi(\cdot)$ is a cumulative distribution function of the standard normal distribution.
Let $\phi(\cdot)$ is a probability density function of the standard normal distribution.
Then, we have
\begin{align*}
f(\mu)
=f_1+f_2 \mu+f_3 \mu^3+O(\mu^3),
\end{align*}
where
\begin{align*}
f_1
&=\frac{\Phi(x/\sigma)}{\Phi(y/\sigma)},
~~~~~
f_2
=-\frac{1}{\sigma}\frac{\phi(x/\sigma)}{\Phi(y/\sigma)}+\frac{1}{\sigma}\frac{\Phi(x/\sigma)\phi(y/\sigma)}{\Phi(y/\sigma)^2}
\intertext{and}
f_3
&=-\frac{1}{\sigma^2}\frac{\phi(x/\sigma)}{\Phi(y/\sigma)}\left(\frac{x}{\sigma}+2\frac{\phi(y/\sigma)}{\Phi(y/\sigma)}\right)
+\frac{1}{\sigma^2}\frac{\Phi(x/\sigma)\phi(y/\sigma)}{\Phi(y/\sigma)}\left(\frac{y}{\sigma}+2\frac{\phi(y/\sigma)}{\Phi(y/\sigma)}\right).
\end{align*}
\end{lemm}

\begin{lemm}
\label{lem6}
Let  $|x|<1$, then
\begin{align*}
&\sum_{n=0}^\infty x^n
=\frac{1}{1-x},
~~~
\sum_{n=0}^\infty nx^{n-1}
=\frac{1}{(1-x)^2}
~~~
\text{and}
~~~
\sum_{n=0}^\infty n(n-1)x^{n-2}
=\frac{2}{(1-x)^3}.
\end{align*}
\end{lemm}

\begin{lemm}
\label{lem7}
Let $w_\alpha$ be a positive value such that 
\begin{align*}
\phi(w_\alpha/v)
=\phi(U/v)-\alpha (\phi(U/v)-\phi(L/v)).
\end{align*}
Then, $w_\alpha\leq z_\alpha$ for all $\alpha\in[0,1]$.
\end{lemm}

Let us prove the theorem.
By noting that the definition of the power, we have
\begin{align}
\text{P}\left(\hat{\theta}_k(t)> z_\alpha\mid \cE \right) 
&= \frac{\Phi((U-\mu)/v)- \Phi((z_\alpha-\mu)/v)}{\Phi((U-\mu)/v)- \Phi((L-\mu)/v)} \nonumber \\
\label{eq;infinite-series}
&=\left\{1-\frac{\Phi((z_\alpha-\mu)/v)}{\Phi((U-\mu)/v)}\right\}\sum_{i=0}^\infty\left\{\frac{\Phi((L-\mu)/v)}{\Phi((U-\mu)/v)}\right\}^i,
\end{align}
almost surely, since $0\leq \Phi((L-\mu)/v)< \Phi((U-\mu)/v)\leq 1$.

By using Theorem \ref{thm2}, $z_\alpha$ can be evaluated by
\begin{align*}
z_\alpha
=v\Phi^{-1}(\Phi(U/v)-\alpha(\Phi(U/v)-\Phi(L/v))),
\end{align*}
and we have
\begin{align*}
\Phi(z_\alpha/v)
=\Phi(U/v)-\alpha(\Phi(U/v)-\Phi(L/v)).
\end{align*}
In addition, by differentiating both side as a function of $v$, we also have
\begin{align}
\label{eq;derivative}
z_\alpha \phi(z_\alpha/v)
=U\phi(U/v)-\alpha(U\phi(U/v)-L\phi(L/v)).
\end{align}
Lemma \ref{lem5} implies that
\begin{align*}
1-\frac{\Phi((z_\alpha-\mu)/v)}{\Phi((U-\mu)/v)}
=1-f_1-f_2\mu-\frac{f_3}{2}\mu^2+O(\mu^3),
\end{align*}
where
\begin{align*}
f_1
&=\frac{\Phi(z_\alpha/v)}{\Phi(U/v)}
=1-\alpha\frac{\Phi(U/v)-\Phi(L/v)}{\Phi(U/v)}, \\
f_2
&=-\frac{1}{v}\frac{\phi(z_\alpha/v)}{\Phi(U/v)}+\frac{1}{v}\frac{\Phi(z_\alpha/v)\phi(U/v)}{\Phi(U/v)^2} \\
&=-\frac{1}{v}\frac{\phi(z_\alpha/v)}{\Phi(U/v)}
+\frac{1}{v}\frac{\phi(U/v)}{\Phi(U/v)}\left(1-\alpha\frac{\Phi(U/v)-\Phi(L/v)}{\Phi(U/v)}\right) \\
&=\frac{1}{v}\frac{\phi(U/v)-\phi(z_\alpha/v)}{\Phi(U/v)}
-\frac{\alpha}{v}\frac{\phi(U/v)}{\Phi(U/v)}\frac{\Phi(U/v)-\Phi(L/v)}{\Phi(U/v)}
\intertext{and}
f_3
&=-\frac{1}{v^2}\frac{\phi(z_\alpha/v)}{\Phi(U/v)}\left(\frac{z_\alpha}{v}+2\frac{\phi(U/v)}{\Phi(U/v)}\right)
+\frac{1}{v^2}\frac{\Phi(z_\alpha/v)\phi(U/v)}{\Phi(U/v)^2}\left(\frac{U}{v}+2\frac{\phi(U/v)}{\Phi(U/v)}\right) \\
&=-\frac{1}{v^2}\frac{\phi(z_\alpha/v)}{\Phi(U/v)}\left(\frac{z_\alpha}{v}+2\frac{\phi(U/v)}{\Phi(U/v)}\right)
+\frac{1}{v^2}\frac{\phi(U/v)}{\Phi(U/v)}\left(\frac{U}{v}+2\frac{\phi(U/v)}{\Phi(U/v)}\right)\left(1-\alpha\frac{\Phi(U/v)-\Phi(L/v)}{\Phi(U/v)}\right).
\end{align*}
On the other hand, we have
\begin{align*}
\frac{\Phi((L-\mu)/v)}{\Phi((U-\mu)/v)}
=g_1+g_2\mu+\frac{g_3}{2}\mu^2+O(\mu^3),
\end{align*}
where 
\begin{align*}
g_1
&=\frac{\Phi(L/v)}{\Phi(U/v)}, \\
g_2
&=-\frac{1}{v}\frac{\phi(L/v)}{\Phi(U/v)}+\frac{1}{v}\frac{\Phi(L/v)\phi(U/v)}{\Phi(U/v)^2}
\intertext{and}
g_3
&=-\frac{1}{v^2}\frac{\phi(L/v)}{\Phi(U/v)}\left(\frac{L}{v}+2\frac{\phi(U/v)}{\Phi(U/v)}\right)
+\frac{1}{v^2}\frac{\Phi(L/v)\phi(U/v)}{\Phi(U/v)^2}\left(\frac{U}{v}+2\frac{\phi(U/v)}{\Phi(U/v)}\right).
\end{align*}
Hence, by applying Lemma \ref{lem6}, the infinite series in (\ref{eq;infinite-series}) can be reduced to
\begin{align*}
\sum_{i=0}^\infty\left\{\frac{\Phi((L-\mu)/v)}{\Phi((U-\mu)/v)}\right\}^i
&=\sum_{i=0}^\infty \left(g_1+g_2\mu+\frac{g_3}{2}\mu^2\right)^{i}+O(\mu^3) \\
&=\sum_{i=0}^\infty g_1^i
+\left(g_2\mu+\frac{g_3}{2}\mu^2\right)\sum_{i=1}^\infty i g_1^{i-1}
+\frac{1}{2}\left(g_2\mu+\frac{g_3}{2}\mu^2\right)^2\sum_{i=1}^\infty i(i-1) g_1^{i-2}
+O(\mu^3) \\
&=\frac{1}{1-g_1}+\frac{g_2}{(1-g_1)^2}\mu
+\left\{\frac{g_3}{2(1-g_1)^2}+\frac{g_2^2}{(1-g_1)^3}\right\}\mu^2+O(\mu^3).
\end{align*}
Denoting $h_1=1/(1-g_1),h_2=g_2/(1-g_2)^2$ and $h_3=g_3/\{2(1-g_1)^2\}+g_2^2/(1-g_1)^3$, we see that
\begin{align*}
h_1
&=\frac{\Phi(U/v)}{\Phi(U/v)-\Phi(L/v)}, \\
h_2
&=\frac{1}{v}\frac{\Phi(L/v)\phi(U/v)-\phi(L/v)\Phi(L/v)}{(\Phi(U/v)-\Phi(L/v))^2},
\intertext{and}
h_3
&=-\frac{1}{2v^3}\frac{L\phi(L/v)\Phi(U/v)-U\Phi(L/v)\phi(U/v)}{(\Phi(U/v)-\Phi(L/v))^2} \\
&-\frac{1}{v^2}\frac{\phi(L/v)\Phi(U/v)-\Phi(L/v)\phi(U/v)}{(\Phi(U/v)-\Phi(L/v))^2}\frac{\phi(U/v)-\phi(L/v)}{\Phi(U/v)-\Phi(L/v)}
\end{align*}
Therefore, we have
\begin{align}
\text{P}\left(\theta_k(t)> z_\alpha\mid (\hat{t},\hat{k})=(t,k)\right)
&=(1-f_1-f_2\mu+f_3\mu^2)(h_1+h_2\mu+h_3\mu^2)+O(\mu^3) \nonumber \\
&=(1-f_1)h_1+\{(1-f_1)h_2-f_2h_1\}\mu \nonumber \\
\label{eq;power}
&+\{(1-f_1)h_3-f_2h_2+f_3h_1\}\mu^2+O(\mu^3).
\end{align}
Moreover, intercept and coefficients in (\ref{eq;power}) can be directly calculated by
\begin{align*}
&(1-f_1)h_1
=\alpha,
~~~
(1-f_1)h_2-f_2h_1
=\frac{\kappa}{v},
\intertext{and}
&(1-f_1)h_3-f_2h_2+f_3h_1
=\frac{\kappa}{v^2}\frac{\phi(U/v)-\phi(L/v)}{\Phi(U/v)-\Phi(L/v)}.
\end{align*}
For the last equality, we just use (\ref{eq;derivative}).

Note that $\phi(x)$ is decreasing when $x>0$.
Lemma \ref{lem7} implies
\begin{align*}
\phi(z_\alpha/v)
&\leq \phi(w_\alpha/v) \\
&=\phi(U/v)-\alpha(\phi(U/v)-\phi(L/v)),
\end{align*}
and it follows that $\kappa\leq 0$.
Finally, by optimizing the leading term in (\ref{eq;power}) with respect to $\mu$, we obtain
\begin{align*}
{\rm P}\left(\theta_k(t)> z_\alpha\mid (\hat{t},\hat{k})=(t,k)\right)
&=\alpha+\frac{\kappa}{v} \mu+\frac{\kappa}{v^2}\frac{\phi(U/v)-\phi(L/v)}{\Phi(U/v)-\Phi(L/v)} \mu^2+O(\mu^3) \\
&\geq \alpha-\frac{\kappa}{4}\frac{\Phi(U/v)-\Phi(L/v)}{\phi(U/v)-\phi(L/v)}+O(\mu^3) \\
&= \frac{3}{4}\alpha+ \frac{1}{4}\frac{\phi(z_\alpha/v)-\phi(U/v)}{\phi(L/v)-\phi(U/v)}+O(\mu^3),
\end{align*}
and this complete the proof.

\section{Proof of Lemmas}
\label{sec:proof2}
In this section, we give the proofs of additional lemmas.
\subsection{Proof of Lemma \ref{lem4}}
\label{subsec:proof-lem4}
Assume that $x_i<x_j$ for some $(i,j)\in I\times I^c$.
It is enough to show that
\begin{align}
\label{eq:ineq}
x_{i_1}+\cdots+x_{i_K}< x_{j_1}+\cdots+x_{j_K}
\end{align}
for some $(j_1,\ldots,j_K)$ such that $j_1<\cdots<j_K$.
Without loss of generality, let $i=i_1$ and $j=j_1$.
Then, (\ref{eq:ineq}) holds by taking $j_k=i_k$ for $k=2,\ldots,K$, and thus $B^c\subset A^c$ is true.
$A\subset B$ is trivial.

\subsection{Proof of Lemma \ref{lem5}}
\label{subsec:proof-lem5}
By using Taylor expansion, we know that
\begin{align*}
f(x)
=f(0)+f'(0)x+f''(0)x^2+O(x^3),
\end{align*}
since $f$ is differentiable with respect to $\mu$.
Direct calculations show the result.

\subsection{Proof of Lemma \ref{lem6}}
\label{subsec:proof-lem6}
Because $|x|<1$, partial sum $\sum_{n=1}^M x^n$ converges to $1/(1-x)$ as $M$ goes to infinity.
Thus by applying term-by-term differentiation, we have the result.

\subsection{Proof of Lemma \ref{lem7}}
\label{subsec:proof-lem7}
Define
\begin{align*}
g(\alpha)
&=\Phi(z_\alpha/v)-\Phi(w_\alpha/v) \\
&=\alpha\Phi(L/v)+(1-\alpha)\Phi(U/v)-\Phi(w_\alpha/v).
\end{align*}
To prove the lemma, we show $g(\alpha)\geq 0$, because this implies the desired result from the monotonicity of $\Phi(\cdot)$.
By the definition of $w_\alpha$, we see that
\begin{align*}
\frac{\text{d}w_\alpha}{\text{d}\alpha}
=-\frac{v^2}{w_\alpha}(\phi(L/v)-\phi(U/v)).
\end{align*}
In addition, since $\phi$ is continuous and $\phi(w_\alpha/v)\to \phi(U/v)$ as $\alpha\to 0$, it must be satisfied that $w_\alpha\to U$ as $\alpha\to 0$.
Since
\begin{align*}
g'(\alpha)
&=\Phi(L/v)-\Phi(U/v)-\phi(w_\alpha/v)\frac{\text{d}w_\alpha}{\text{d}\alpha} \\
&=\Phi(L/v)-\Phi(U/v)+v^2(\phi(L/v)-\phi(U/v))\frac{\phi(w_\alpha/v)}{w_\alpha}
\end{align*}
is non-decreasing on $\alpha\in[0,1]$, we have
\begin{align*}
g'(\alpha)
&\geq \lim_{\alpha\to0}g'(\alpha) \\
&=\Phi(L/v)-\Phi(U/v)
+\frac{v^2}{U}\phi(U/v)(\phi(L/v)-\phi(U/v)).
\end{align*}
Here, the monotonicity of $g'(\alpha)$ can be easily verified by considering $g''(\alpha)$.
By noting that the right-hand side in the above inequality is also non-decreasing as a function with respect to $L$, we see that $g'(\alpha)$ can be bounded below as
\begin{align*}
g'(\alpha)
\geq \lim_{L\to \infty}\Biggl\{\Phi(L/v)-\Phi(U/v)
+\frac{v^2}{U}\phi(U/v)(\phi(L/v)-\phi(U/v))\Biggr\}
=0,
\end{align*}
where we use $U\to\infty$ as $L\to \infty$ in the last equality.
Therefore, $g(\alpha)$ is non-decreasing with respect to $\alpha$.
Finally, by taking limit as $\alpha\to 0$ in $g(\alpha)$, we obtain the result.


\end{document}